\documentclass{article}

 \usepackage[preprint]{neurips_2025}

\usepackage[utf8]{inputenc} 
\usepackage[T1]{fontenc}    
\usepackage{hyperref}       
\usepackage{url}            
\usepackage{booktabs}       
\usepackage{amsfonts}       
\usepackage{nicefrac}       
\usepackage{microtype}      
\usepackage{xcolor}         
\usepackage{makecell}
\usepackage[colorinlistoftodos,prependcaption,textsize=small]{todonotes}
\usepackage{graphicx}
\usepackage{caption}    

\usepackage{amsmath,amsfonts,bm}

\def\eqref#1{equation~\ref{#1}}

\def\1{\bm{1}}

\DeclareMathAlphabet{\mathsfit}{\encodingdefault}{\sfdefault}{m}{sl}
\SetMathAlphabet{\mathsfit}{bold}{\encodingdefault}{\sfdefault}{bx}{n}

\newcommand{\calD}{\mathcal{D}}

\newcommand{\bydef}{\stackrel{\text{def}}{=}}

\usepackage{algorithm}
\usepackage{algpseudocode}

\usepackage{multirow}
\usepackage{enumitem}
\usepackage[most]{tcolorbox}
\usepackage{float}
\usepackage{caption}
\usepackage{subcaption}
\usepackage{longtable}
\usepackage{listings}
\usepackage{amssymb}
\usepackage{mathtools}
\usepackage{amsthm}
\usepackage{multirow}
\usepackage{multicol}
\usepackage{longtable}
\usepackage{titletoc}

\newcommand{\red}[1]{\textcolor{red}{#1}}

\newcommand{\teal}[1]{\textcolor{teal}{#1}}

\title{MedGUIDE: Benchmarking Clinical Decision-Making in Large Language Models}

\author{%
    Xiaomin Li\thanks{Co-first authors: Xiaomin Li and Mingye Gao. Correspondence to Xiaomin Li \texttt{(xiaominli@g.harvard.edu)}.} \\
  Harvard University\\
  \And
  Mingye Gao \\
  MIT\\
  \And
  Yuexing Hao \\
  Cornell University\\ MIT\\ Mayo Clinic \\
  \AND
  Taoran Li \\
  UIUC \\
  \And
  Guangya Wan \\
  University of Virginia \\
  \And
  Zihan Wang \\
  Harvard Medical School \\
  Abaka AI\\
  \And
  Yijun Wang \\
  Harvard University \\
  \And
  Xupeng Chen \\
  NYU\\
}

\begin{document}

\maketitle
\begin{abstract}
Clinical guidelines, typically structured as decision trees, are central to evidence-based medical practice and critical for ensuring safe and accurate diagnostic decision-making. However, it remains unclear whether Large Language Models (LLMs) can reliably follow such structured protocols. In this work, we introduce \textbf{MedGUIDE}, a new benchmark for evaluating LLMs on their ability to make guideline-consistent clinical decisions. MedGUIDE is constructed from 55 curated NCCN decision trees across 17 cancer types and uses clinical scenarios generated by LLMs to create a large pool of multiple-choice diagnostic questions. We apply a two-stage quality selection process, combining expert-labeled reward models and LLM-as-a-judge ensembles across ten clinical and linguistic criteria, to select 7,747 high-quality samples. We evaluate 25 LLMs spanning general-purpose, open-source, and medically specialized models, and find that even domain-specific LLMs often underperform on tasks requiring structured guideline adherence. We also test whether performance can be improved via in-context guideline inclusion or continued pretraining. Our findings underscore the importance of MedGUIDE in assessing whether LLMs can operate safely within the procedural frameworks expected in real-world clinical settings.
\end{abstract}

\section{Introduction}

Clinical guidelines play a critical role in modern medicine, providing structured recommendations for diagnosis and treatment that are informed by expert consensus and current evidence. These guidelines, such as those published by the National Comprehensive Cancer Network (NCCN) or the American Heart Association (AHA), are often encoded as decision trees, offering standardized pathways for handling diverse patient scenarios \citep{nccn_guidelines, arnett2019acc}. Adherence to these protocols is critical for ensuring consistent, safe, and high-quality clinical decision-making.

Large Language Models (LLMs) have demonstrated remarkable capabilities across general and medical natural language processing (NLP) tasks, including biomedical question-answering (QA) \citep{lee2020biobert, alsentzer2019publicly, shin2020biomegatron}, clinical dialogue modeling \citep{sun2024llamacare, he2024bp4er}, and even diagnostic reasoning \citep{singhal2025toward, yang2022gatortron, wu2023pmc}. However, most evaluations to date focus on factual recall or general in-context reasoning, rather than testing whether LLMs can follow domain-specific decision rules as clinicians must in practice. Emerging work has begun to explore instruction and rule-following behavior in LLMs \citep{zheng2023judging, dong2024can, rulebench_arxiv}, yet few studies directly assess LLM adherence to formal clinical pathways, particularly in complex diagnostic contexts \citep{fast2024autonomous, huang2024comprehensivesurveyevaluatinglarge}.

To fill this research gap, we introduce \textbf{MedGUIDE}—\textbf{G}uideline \textbf{U}nderstanding and \textbf{I}nference for \textbf{D}ecision \textbf{E}valuation—a benchmark designed to evaluate whether LLMs can make diagnostic decisions in accordance with established medical guidelines. MedGUIDE is constructed from 55 decision trees curated from NCCN oncology protocols, covering 17 of the most common cancer types. We transform these trees into clinical vignettes and corresponding multiple-choice questions (MCQs) that require selecting the correct next step in a patient’s management plan. Rather than testing general knowledge alone, MedGUIDE probes whether models can apply structured clinical logic. We implement a rigorous and efficient two-stage filtering process using both human-annotated reward models \citep{glaese2022improving, wang2024helpsteer2} and LLM-as-a-judge ensembles \citep{chen2024mllm, huang2024empirical, polo2024efficient} to ensure the final dataset is both clinically plausible and textually well-formed. From a raw pool of 16,000 QA samples, we retain 7,747 high-quality examples.

We evaluate 25 LLMs, including general-purpose, open-source, and medically fine-tuned models, using both standard accuracy and a weighted accuracy metric that accounts for question difficulty. To better understand the LLM's capabilities tested by MedGUIDE, we compare model performance on MedGUIDE against other established benchmarks, including IFEval \citep{dong2024can} and MMLU-Professional Medicine \citep{hendrycks2020measuring}, and analyze cross-benchmark correlations. In addition, we explore two methods for further improving the model's capability of adhering to the clinical guideline: (1) \textit{guideline-in-context prompting}, which supplies the model with the relevant decision tree during inference; and (2) \textit{continued pretraining on guidelines}, which aims to internalize the structure and logic of clinical pathways and apply them to diagnostic or treatment planning tasks in MedGUIDE.

\noindent\textbf{Our key contributions are:}
\begin{itemize}
    \item We introduce \textbf{MedGUIDE}, the first benchmark focused on evaluating LLMs' ability to follow structured clinical decision trees based on real-world medical guidelines. \footnote{MedGUIDE dataset: \url{https://huggingface.co/datasets/MedGUIDE/MedGUIDE-MCQA-8K}. Code for experiments: \url{https://anonymous.4open.science/r/Submission-MedGUIDE-187A}}
    \item We construct a high-quality multiple-choice questions (MCQs) dataset using 55 NCCN decision trees, and apply a dual-stage-filtering pipeline combining expert-labeled reward models and ensemble LLM scoring.
    \item We benchmark 25 diverse LLMs and reveal significant limitations in both general and medical models’ ability to align with guideline-based decision logic.
    \item We analyze correlations with existing benchmarks and show that MedGUIDE evaluates capabilities beyond factual recall, including structured guideline comprehension and task-specific adherence.
    \item We evaluate whether guideline grounding, via in-context prompting or continued pertaining, improves LLM performance on MedGUIDE.
\end{itemize}

\begin{figure}[htbp]
  \centering
  \includegraphics[width=\textwidth]{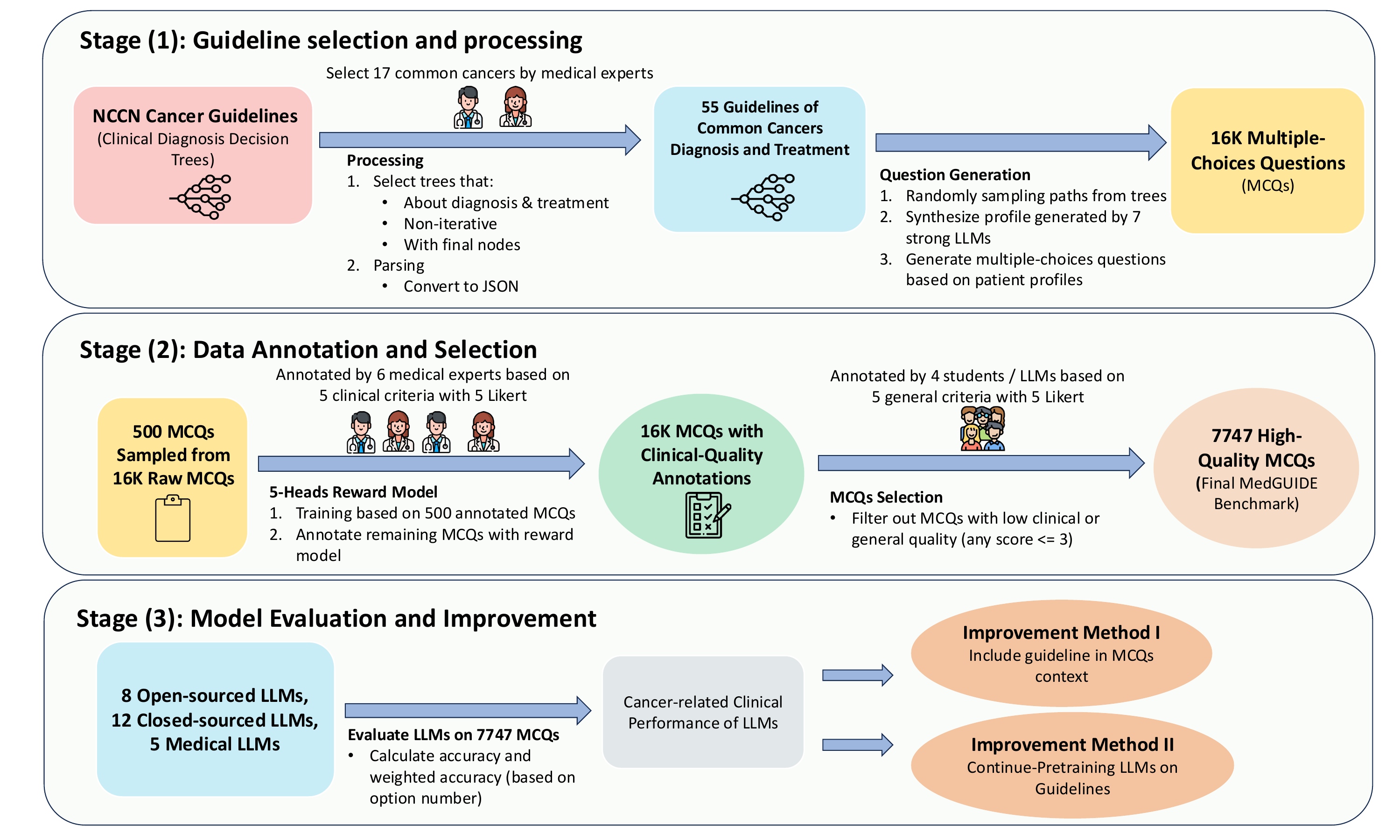}
  \caption{Overview of the MedGUIDE Benchmark Pipeline. \textbf{Stage (1)}: selecting and processing 55 NCCN clinical decision tree guidelines for 17 common cancers to generate 16K synthetic MCQs. \textbf{Stage (2)}: annotating and filtering these MCQs based on clinical and general quality criteria using expert-labeled data and a 5-head reward model, resulting in a curated set of 7,747 high-quality MCQs. \textbf{Stage (3)}: evaluating 25 LLMs (open-source, closed-source, and medical) on the benchmark and applying two improvement methods—guideline-aware prompting (Method I) and guideline-based continued pretraining (Method II).
  }
  \label{Study_Design}
\end{figure}

\section{Related Work}\label{sec:RelatedWork} 

\textbf{Medical LLMs for Clinical Diagnosis.}\quad
The application of Large Language Models (LLMs) in healthcare has progressed from early biomedical pretraining \citep{lee2020biobert, alsentzer2019publicly, shin2020biomegatron, wang2023clinicalGPT} to instruction-tuned and dialogue-optimized systems \citep{singhal2025toward, yang2022gatortron, wu2023pmc, venigalla2022biomedlm} that support medical reasoning tasks. Recent efforts further expand capabilities via: (1) retrieval-augmented generation using medical corpora \citep{wen2023mindmap, wu2024guiding, shi2023retrieval, ranjit2023retrieval, ge2023development}, databases \citep{shi2023retrieval, rau2023context, wang2024beyond, chen2024large}, and knowledge graphs \citep{wen2023mindmap, wu2024guiding, gao2023large, zhu2024emerge}; (2) supervised fine-tuning on clinical notes, dialogues, and multimodal inputs \citep{toma2023clinical, jiang2023health, sun2024llamacare, wu2024medkp}; and (3) RLHF strategies, both online \citep{wang2023clinicalGPT, zhang2023huatuoGPT, zhou2024large} and offline \citep{dou2024integrating, yang2024pediatricsGPT}.

\textbf{Medical Benchmarks for LLMs.}\quad
LLM benchmarks in medicine span from knowledge-focused tasks \citep{pubmedqa_duke, medmcqa} to more complex clinical reasoning and scenario-based evaluations \citep{blurmb_medrxiv, consumer_med_qa_arxiv, clinical_calc_qa, multi_lingual_med_exam_qa}. A growing body of work emphasizes guideline adherence \citep{clinical_consent_arxiv, proai_arxiv}, yet few benchmarks systematically test whether LLMs can follow structured protocols such as those from NCCN \citep{nccn_guidelines} or AHA \citep{arnett2019acc}. Prior work addressing guideline-following remains limited in scope \citep{fast2024autonomous}, motivating our MedGUIDE benchmark, which uniquely evaluates both diagnostic accuracy and fidelity to established clinical pathways.

\textbf{Instruction and Guidance Following.}\quad
Instruction following is a fundamental competency of LLMs \citep{instruction_tuning_ibm}, while guidance following requires models to adhere to structured, domain-specific rules \citep{guidance_who}, such as clinical practice guidelines \citep{Hager2024Evaluation}. Evaluating guideline adherence remains challenging, often relying on proxy tasks or human review. Recent benchmarks study rule-following under varying task structures \citep{chen2024mllm, huang2024empirical, weyssow2024codeultrafeedback, polo2024efficient, infobench_acl, rulebench_arxiv, llmbar_arxiv}. Our work extends this line by examining how well LLMs follow multistep clinical decision trees under real-world constraints.

\textbf{Reward Models and LLM-as-a-Judge.}\quad
Prior work has employed multi‐head reward models to score generated outputs along several attributes during post‐training stages \citep{glaese2022improving, wang2024helpsteer2, li2024rule, li2025data, wang2024interpretable, li2025multi}, an approach we adopt for our dataset’s quality‐based filtering.  Complementing this, \textit{LLM‐as‐a‐judge} methods directly prompt a large model to rate sample quality, which has proven effective for large‐scale evaluation and bias reduction \citep{zheng2023judging, li2025data, chen2024mllm, huang2024empirical, polo2024efficient, dong2024can, thakur2024judging}. 
\section{MedGUIDE: A Guideline-Based Clinical Decision-Making Benchmark}
\label{sec:Dataset}

\subsection{NCCN Guidelines for Cancer}\label{subsec:Dataset-NCCN}

The National Comprehensive Cancer Network (NCCN) guidelines \citep{nccn_guidelines} are comprehensive, regularly updated protocols for cancer care—spanning prevention, diagnosis, treatment, and supportive care—and are organized as decision trees, with root nodes capturing key clinical variables (e.g., symptoms, labs, history) and branches splitting according to defined criteria or thresholds.  

For example, Figure~\ref{fig:ExampleDecisionTree} illustrates a decision tree for first relapse in acute promyelocytic leukemia (APL), distinguishing among early relapse with or without arsenic trioxide exposure and late relapse. Because these pathways are designed for general clinical applicability rather than individualized care, they provide broad management strategies rather than patient-specific recommendations. The guidelines are published in PDF format behind a login portal, which limits direct ingestion by LLMs. To construct MedGUIDE, two medical experts selected 17 common cancers (listed in Table~\ref{tab:disease-abbreviation} in the Appendix) and their associated diagnostic decision trees. We excluded trees with loops or ambiguous leaf nodes, resulting in 55 well-structured guideline trees used for generating synthetic multiple-choice questions (MCQs).

\subsection{Synthetic Prompt Generation}
\label{subsec:Dataset-GeneratePrompts}
We convert each decision tree into JSON format and use GPT-4o to enumerate all valid paths from the root to each leaf. These paths are manually reviewed to ensure correctness. Each path corresponds to a plausible clinical scenario that matches the criteria described along its trajectory. For example, in the case illustrated in Figure~\ref{fig:ExampleDecisionTree}, the patient follows the highlighted path because he experienced a \textbf{first relapse} five months after completing treatment with \textbf{ATRA and arsenic trioxide}---qualifying as an \textbf{early relapse (<6 mo)} after ATRA without anthracycline exposure. As a result, the recommended regimen is \textbf{arsenic trioxide $\pm$ ATRA $\pm$ gemtuzumab ozogamicin}, and if a \textbf{second morphologic remission} is achieved and the patient is \textbf{not a transplant candidate}, the guideline suggests \textbf{arsenic trioxide consolidation (total of 6 cycles)} as the next step.

Given a path (excluding the leaf), we prompt multiple LLMs to generate a clinical profile consistent with the scenario. We use the following models: \texttt{GPT-4o-mini} \citep{openai2024gpt4omini}, \texttt{GPT-4o} \citep{openai2024gpt4o}, \texttt{GPT-4.1} \citep{openai2025gpt41}, \texttt{Claude-3.5-Haiku} \citep{anthropic2024computer}, \texttt{Claude-3.7-Sonnet} \citep{anthropic2025claude}, \texttt{DeepSeek-V3} \citep{liu2024deepseek}, \texttt{Gemini-2.5-Flash} \citep{google2025geminiflash}, \texttt{Llama-3.2-1B-Instruct} \citep{meta2024llama32}, \texttt{Llama-3.1-Instruct} (8B, 70B) \citep{meta2024llama31}, \texttt{Qwen2.5-Instruct} (7B, 32B)  \citep{yang2024qwen2}, \texttt{Mistral-7B-Instruct} \citep{jiang2023mistral7b}, and \texttt{Mixtral-8x7B-Instruct} \citep{jiang2024mixtral}. This model diversity helps ensure variation in language and scenario framing.

Each profile is converted into a multiple-choice QA format by appending a question about the appropriate next clinical step. The answer options include all possible leaf nodes from the same decision tree, with the correct answer corresponding to the actual leaf node on the source path. This process yields a total of 16,000 multiple-choice QA pairs.

\begin{tcolorbox}[colback=blue!3!white, colframe=blue!50!white, title=Example of QA Data, left=10mm, breakable]
\small 

\begin{center}
  \includegraphics[width=0.8\linewidth]{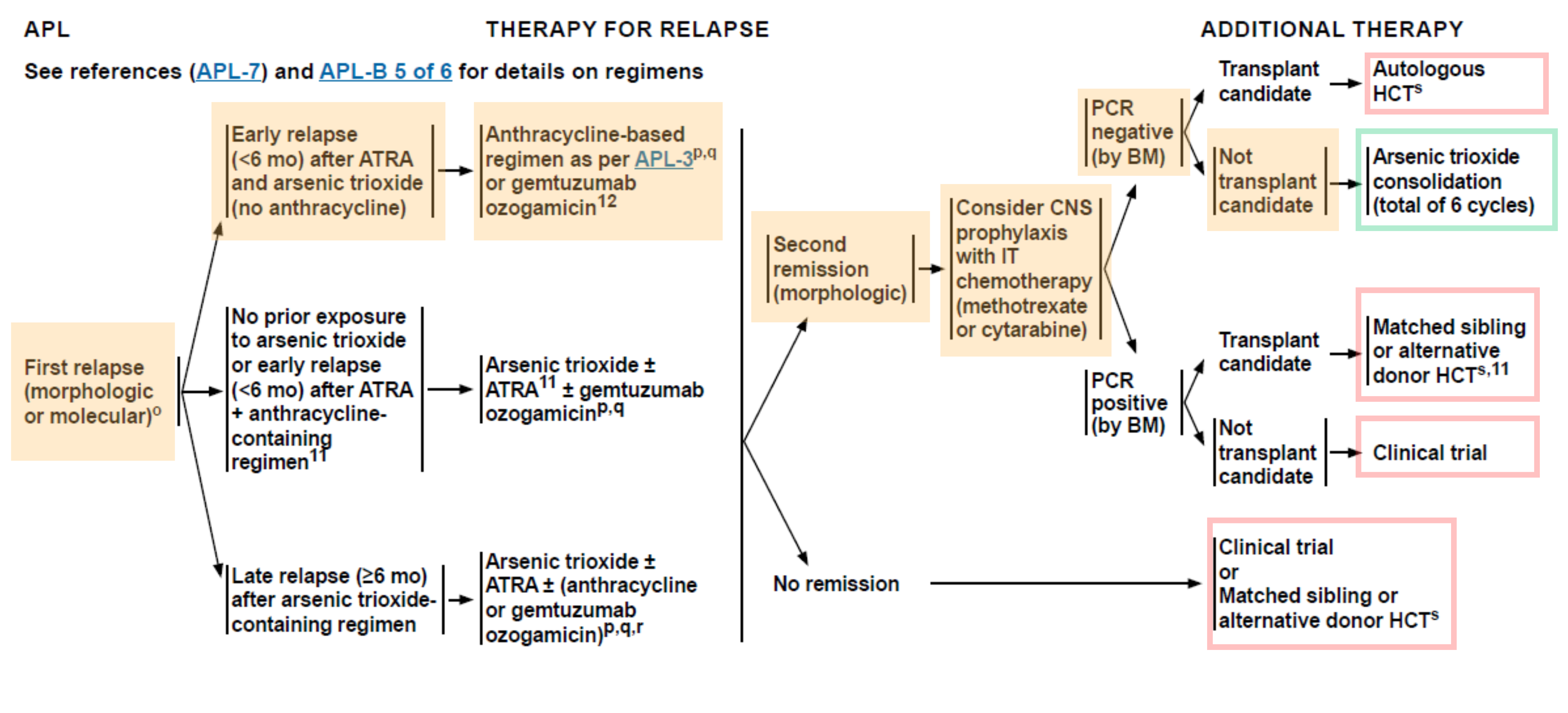}
  \captionof{figure}{\textbf{NCCN Acute Myeloid Leukemia (AML) Guideline.} The orange boxes illustrate the workflow through which the sample QA dataset is generated. Red and green annotations represent the correct and incorrect options. }\label{fig:ExampleDecisionTree}
\end{center}

\textbf{Prompt:} A 32-year-old male patient with acute promyelocytic leukemia (APL) was initially diagnosed one year ago and achieved complete remission after treatment with all-trans retinoic acid (ATRA) and arsenic trioxide. He experienced his first relapse five months after completing these treatments, with both morphologic and molecular evidence of disease recurrence. He has no prior exposure to anthracycline therapy, making him eligible for a more aggressive treatment approach. Upon reevaluation, he received an anthracycline-based regimen according to the APL-3 protocol, resulting in a second morphologic remission confirmed by bone marrow biopsy. Given his potential central nervous system involvement and current PCR negativity in bone marrow, the team is contemplating prophylactic measures. He is not a transplant candidate due to comorbidities and overall health status. As consolidation therapy, the plan is to initiate arsenic trioxide for a total of six cycles. Given this clinical scenario, what would be the appropriate next step in management for consolidation therapy?

\smallskip
\textbf{Options:}

\begin{enumerate}[label=(\Alph*)]
\item Clinical trial
\item Arsenic trioxide consolidation
\item Clinical trial or matched sibling/alternative donor HCT
\item Matched sibling or alternative donor HCT
\item Autologous HCT
\end{enumerate}
\smallskip
\textbf{Correct Answer:}
(B) Arsenic trioxide consolidation.

\end{tcolorbox}

\subsection{Quality-Based Selection via Reward Models}
\label{subsec:Dataset:QualitySelection}

\subsubsection{Clinical and General Criteria}





To ensure the high quality of our MCQs data, we adopt a quality-based data selection framework. In collaboration with six medical experts, we define five clinically grounded evaluation criteria; meanwhile, we introduce five general quality criteria to evaluate language and structure of MCQs (\ref{tab:criteria}). 
Each criterion is rated on a 5-point Likert scale. Detailed rubrics are provided in Appendix~\ref{sec:Appendix-Criteria}.

\begin{table}[htbp]
\centering
\caption{Clinical and General Evaluation Criteria}
\begin{tabular}{|c|c|}
\hline
\textbf{Clinical Evaluation Criteria} & \textbf{General Quality Criteria} \\
\hline
Clinical Plausibility & Clarity and Detail Level \\
Clinical Utility & Consistency and Internal Logic \\
Quality of Decision Path & Safety and Toxicity \\
Alignment to Decision Path & Textual Quality and Professionalism \\
Clinical Accuracy of Correct Answer & Option Distinctiveness \\
\hline
\end{tabular}
\label{tab:criteria}
\end{table}

\subsubsection{Training Reward Models}
Due to time and resource constraints, we propose an efficient data-annotation pipeline. Six medical experts from Harvard Medical School, MIT Biology, and Harvard Stem Cell Institute (detailed demographic information provided in Appendix~\ref{sec:Appendix-Annotators}) first labeled a randomly-selected subset of 500 MCQs according to the 5 clinical evaluation criteria shown in Table \ref{tab:criteria}; each MCQ is assigned a single score for each criterion,  resulting in a vector $[R_1(x), R_2(x), \dots, R_5(x)] \in \{1, 2, \dots, 5\}^5$. We then trained a 5-head reward model using \texttt{Qwen2.5-7B-Instruct} as the backbone, with 5 epochs and a learning rate of $2 \times 10^{-5}$ (hyperparameter tuning details in Appendix~\ref{sec:Appendix-RewardModelTraining}). The trained model was then used to rate the remaining 15,500 MCQs.

For general quality, we used an ensemble of LLM-as-a-judge models. Specifically, we queried \texttt{GPT-4o-mini}, \texttt{Claude 3.5-Haiku}, \texttt{Gemini 2.5-Flash}, and \texttt{DeepSeek-V3}, and averaged their scores across each criterion. The ensemble methods help reduce rating bias \citep{schoenegger2024wisdom, li2024llms, chen2025harnessing}. To validate these automatic ratings, four human annotators independently reviewed a random subset of 500 samples, achieving over 96\% agreement across all annotators.

\subsubsection{Data Selection}
After obtaining quality scores from the reward models, we apply a filtering step to retain only the highest-quality samples for the final dataset. For each sample $x \in \calD$, we collect ten rating scores covering both clinical and general evaluation criteria, denoted by $\{R_1(x), R_2(x), \dots, R_{10}(x)\}$. To ensure robustness and consistency across all dimensions, we apply two selection conditions: (1) the minimum of the ten scores must be strictly greater than 2, and (2) the average score must exceed 3. This dual criterion ensures that selected samples exhibit both overall strength and no major weaknesses in any specific dimension. Formally, the selection rule is:
\begin{equation}
    \calD^{\ast} = \left\{x: \min_{1 \leq i\leq 10} R_i(x) > 2 \text{ and } \sum_{i=1}^{10} R_i(x) > 3, \quad \text{ for } x \in \calD \right\}
\end{equation}

This filtering step yields 7,747 high-quality samples, which constitute our final MedGUIDE benchmark dataset. The distribution of these samples across various dimensions (such as cancer types, number of answer options, and LLMs used for generation) is illustrated in Figure~\ref{fig:data_distribution} below.

\begin{figure}[ht]
    \centering
    \includegraphics[width=\linewidth]{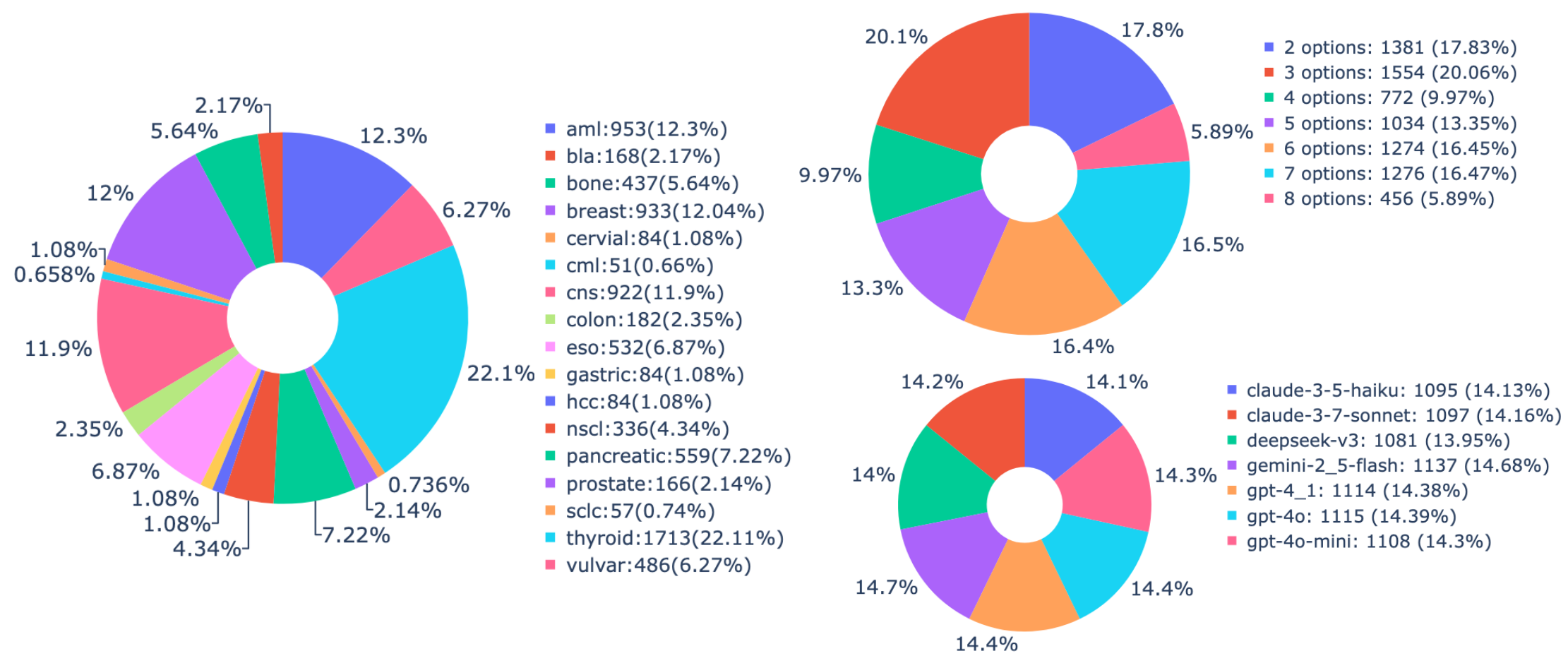}
  \caption{Distributions of cancer types (left), MCQ option counts (top right), and LLMs used for question generation (bottom right).}
  \label{fig:data_distribution}
\end{figure}

\section{Experiments}
We conduct experiments to evaluate the ability of various LLMs to reason about next steps in clinical diagnosis using the MedGUIDE benchmark.

\subsection{Models:}
We evaluated a total of 25 models (at inference temperature 0) spanning a range of sizes and architectures, aiming to ensure broad coverage and diversity for comprehensive assessment. For analysis, we loosely categorize these models into the following groups:
\begin{itemize}
    \item \textbf{General LLMs (closed-source)}: \texttt{GPT-4o-mini}, \texttt{GPT-4.1}, \texttt{O1} and \texttt{O4-mini} \citep{openai2025o3o4mini}, \texttt{Claude-3.5-Haiku}, \texttt{Claude-3.7-Sonnet}, and \texttt{Deepseek} (V3, R1) \citep{liu2024deepseek, guo2025deepseek}.
    \item \textbf{General LLMs (open-source)}: \texttt{Llama-3.2-Instruct} (1B, 3B), \texttt{Llama-3.1-Instruct} (8B, 70B), \texttt{Mistral-7B-Instruct}, \texttt{Mixtral-8x7B-Instruct}, \texttt{Qwen2.5-Instruct} (1.5B, 7B), and \texttt{Qwen3} (4B, 8B, 14B, 32B) \citep{qwen2025qwen3}.
    \item \textbf{Medical LLMs:} \texttt{ClinicalCamel-70B} \citep{toma2023clinical}, \texttt{Medalpaca} (7B, 13B) \citep{han2023medalpaca}, and \texttt{Meditron} (7B, 70B) \citep{chen2023meditron}.
\end{itemize}

\subsection{Evaluation Metrics}
Each MedGUIDE sample is a multiple-choice question (MCQ), making \textit{accuracy} the natural baseline metric. However, since the number of answer options varies across questions, we introduce a \textit{weighted accuracy} metric that accounts for the difficulty of each sample. Let $c(x)$ denote the number of options in sample $x$. We define the difficulty function as 
\begin{equation}
    f(x) \bydef 1-\frac{1}{c(x)},
\end{equation}
which reflects the margin above random guess performance. Then we assign weight to each sample according to:
\begin{equation}\label{eq:WeightedAccuracy-weight}
    w_i \bydef \frac{f(x_i)}{ \sum_{i=1}^{|\calD^\ast|} f(x_i)}
    = \frac{1-\frac{1}{c(x_i)}}{|\calD^\ast| -  \sum_{i=1}^{|\calD^\ast|} \frac{1}{c(x_i)}}.
\end{equation}
Denote $ \mathrm{1}(\cdot)$ as the indicator function. Then the final weighted accuracy is defined as:
\begin{equation}\label{eq:WeightedAccuracy}
    \text{WeightedAccuracy}(\calD^\ast) \bydef \frac{ \sum_{i=1}^{|\calD^\ast|} w_i \cdot \mathrm{1}(\hat{y}_i = y_i)}{|\calD^\ast|},
\end{equation}

\subsection{Results}
\begin{figure}[ht]
    \centering
    \includegraphics[width=0.7\textwidth]{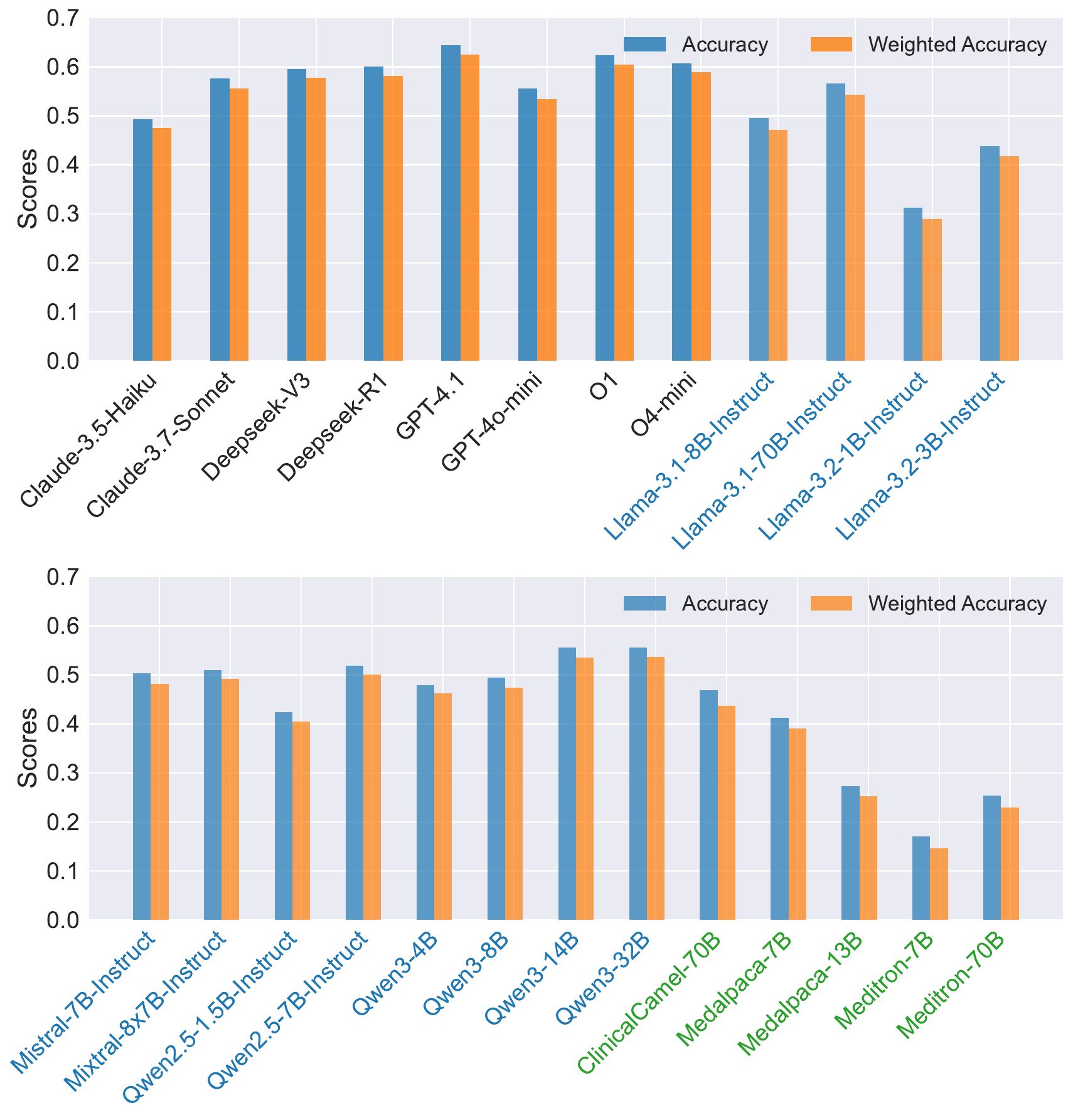}
    \caption{Benchmark evaluation results}
    \label{fig:MainResults}
\end{figure}

The results of all 25 models under both accuracy metrics are plotted in Figure~\ref{fig:MainResults}. We observe that closed-source models, which are generally state-of-the-art across many tasks \citep{achiam2023gpt, openai2024o1, guo2025deepseek, liu2024deepseek, anthropic2025claude}, outperform others. In particular, \texttt{GPT-4.1} achieves the best performance, followed by reasoning-augmented models like \texttt{O1} and \texttt{O4-mini}. While these reasoning models perform well, they incur higher inference costs due to the extra thinking tokens. Among open-source models, we find that increasing model size does not always lead to better performance. For instance, \texttt{Qwen3-14B} and \texttt{Qwen3-32B} do not significantly outperform \texttt{Qwen3-4B}, and \texttt{Mixtral-8x7B} shows no large advantage over \texttt{Mistral-7B}. Notably, the \texttt{Qwen3} series shows strong performance despite smaller sizes.

Surprisingly, all medical LLMs underperform, regardless of size. We hypothesize two reasons: 1. MedGUIDE emphasizes reasoning over clinical decision paths, which requires models to \textbf{apply medical knowledge contextually}. Medical LLMs may possess domain knowledge but lack the reasoning capabilities required for next-step prediction. 2. Most medical LLMs are based on older backbones (e.g., Llama2), which may \textbf{struggle with instruction-following and logical deduction}. These reasoning skills are essential for serving as reliable medical assistants. Thus, MedGUIDE not only evaluates medical knowledge but also tests fundamental LLM capabilities crucial for clinical support tasks.

\subsection{Correlation with Other Benchmarks}

To better understand what capabilities MedGUIDE evaluates, we examine its correlation with other established benchmarks. In particular, we ask: Does performance on MedGUIDE reflect general medical knowledge alone, or does it additionally require broader reasoning or instruction-following abilities? To explore this, we evaluate the models on two complementary benchmarks: \textbf{IFEval}, which targets instruction-following under factual constraints, and \textbf{MMLU-Professional Medicine}, a knowledge-heavy subtask from MMLU that tests factual recall across a wide range of clinical topics. Detailed descriptions of both benchmarks are provided in Appendix~\ref{subsec:Appendix-BenchmarkCorrelation-Details}.

To quantify alignment, we compute three correlation metrics: \textbf{Spearman's $\rho$} \citep{spearman1904general} , \textbf{Kendall's $\tau$} \citep{kendall1938new}, and \textbf{Pearson's $r$} \citep{pearson1895note}, and report full results in Appendix~\ref{subsec:Appendix-BenchmarkCorrelation-Results}. Here, we focus on \textbf{Spearman's $\rho$}, which reflects rank-order agreement between benchmarks. We find that MedGUIDE exhibits strong Spearman correlation with MMLU-Professional Medicine ($\rho = 0.85$), suggesting that medical knowledge remains an important component. However, its nontrivial correlation with IFEval ($\rho = 0.71$) indicates that instruction-following and decision-step reasoning also contribute meaningfully to performance. Together, these results imply that MedGUIDE captures a unique intersection of knowledge recall and structured clinical decision-making not fully represented by either benchmark alone.

\section{Improve Performance}\label{sec:Improve}
\subsection{Method I: Include the Guideline in Context}
\label{subsec:Improve-Method1}

Our first strategy for boosting performance is to provide the relevant guideline decision tree (in JSON form) directly in the model’s prompt. In this case, the model likewise does not need to recall the protocol from scratch but can parse and follow the tree structure in the context. In clinical practice, a physician would readily know which guideline applies. To mimic real‐world usage, we also trained a lightweight Qwen-4B classifier to select the appropriate guideline given the patient vignette, achieving 98\% validation accuracy, demonstrating that identifying the correct guideline is straightforward for both clinicians and capable LLMs.  

\begin{figure}[ht]
    \centering
    \begin{minipage}[b]{0.64\textwidth}
        \centering
        \includegraphics[width=\textwidth]{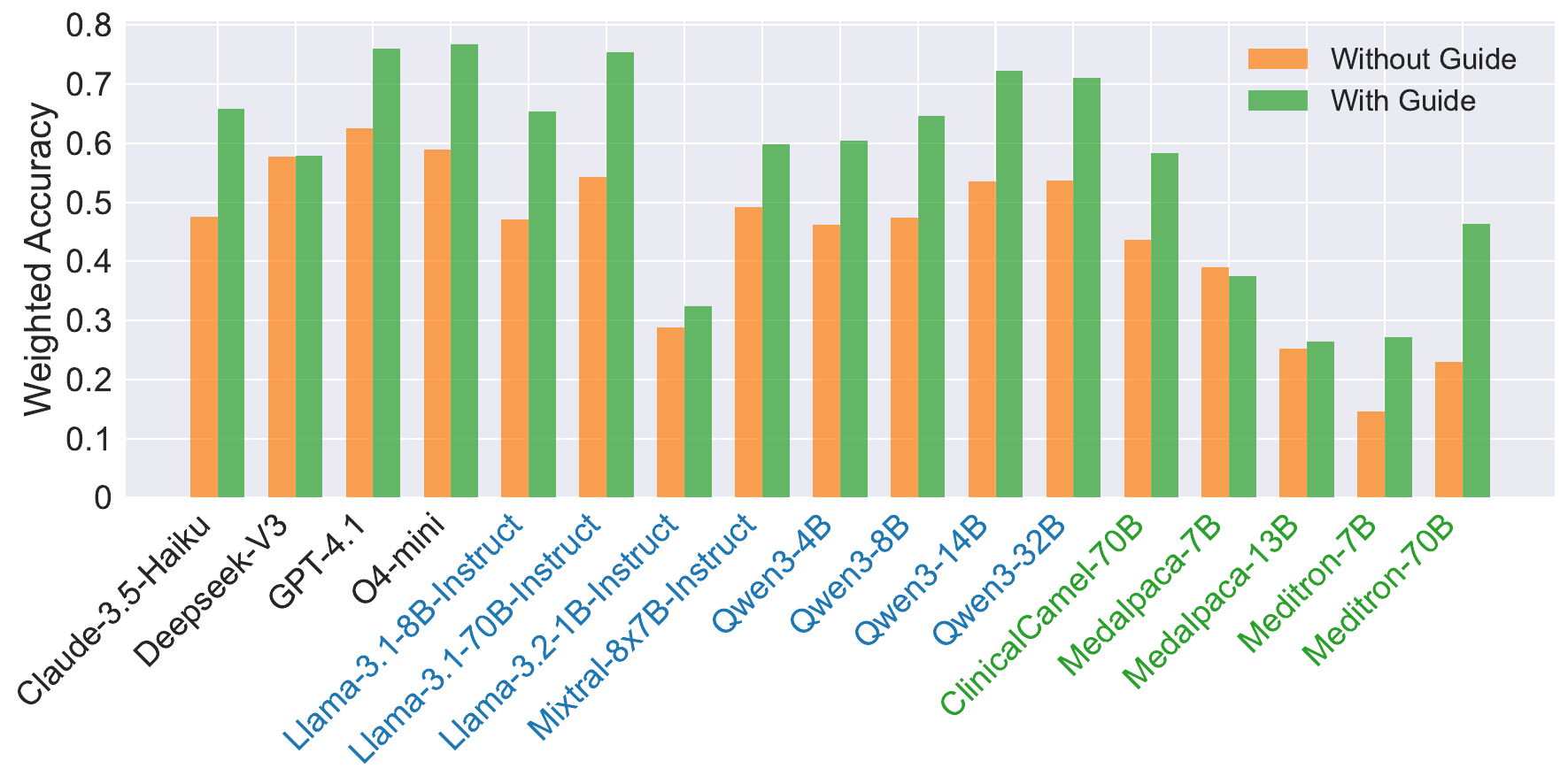}
        \caption{Weighted accuracy before and after using Method I}
        \label{fig:Method1-Improvement}
    \end{minipage}
    \hfill
    \begin{minipage}[b]{0.35\textwidth}
        \centering
        \includegraphics[width=\textwidth]{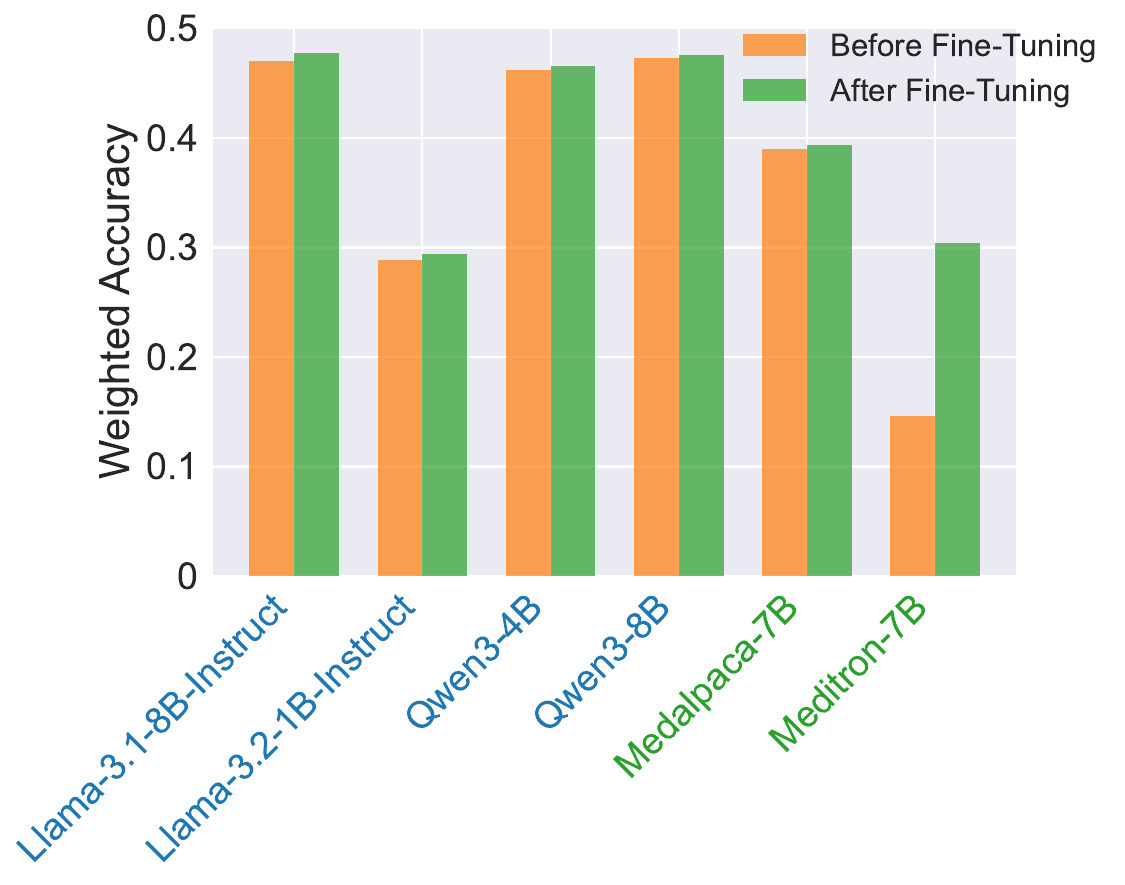}
        \caption{Weighted accuracy before and after using Method II}
        \label{fig:Method2-Improvement}
    \end{minipage}
\end{figure}

\textbf{Results:} Improvements are clear across most models (see Figure~\ref{fig:Method1-Improvement}). For example, Meditron-70B’s weighted accuracy rises from 0.230 to 0.462 (a 102\% relative increase) while Meditron-7B improves by 86\%. Detailed percentage gains for both metrics are shown in Figure~\ref{fig:Method1-ImprovementPercentage} in the Appendix. We also include a detailed case study in Appendix~\ref{subsec:Appendix-Method1}, comparing model outputs before and after applying Method I. In that case study, we ask the model to generate both the answer and an explanation. With the guideline in context, the model selects the correct answer and provides a step-by-step explanation that mirrors the decision-tree logic—from metastatic diagnosis through local control and adjuvant therapy to lung-only complete response and the final recommendation. In contrast, without the guideline, the model chooses an incorrect option and offers a plausible but guideline-inconsistent rationale, overlooking the patient’s complete pulmonary response. This demonstrates the value of including structured guideline information to enhance both answer accuracy and explanation quality.

\subsection{Method II: Fine-tuning with Guideline Data}\label{subsec:Improve-Method2}

In the second approach, we investigate whether a model can internalize clinical guideline knowledge through continued pretraining and subsequently apply it to MedGUIDE without receiving explicit guideline context at inference time.  
Rather than injecting the guideline into the prompt (as in Method I), we perform lightweight fine-tuning using the 55 NCCN decision trees (in structured JSON format).  
This serves as a form of domain-adaptive pretraining, aimed at infusing the model with the logical structure and semantics of the guidelines. We fine-tune a subset of models due to resource constraints, selecting representative general and medical LLMs. Each model is trained for 8 epochs with a learning rate of $1 \times 10^{-5}$ (hyperparameters were selected via a grid search).

\textbf{Results:}
Figure~\ref{fig:Method2-Improvement} and Appendix Figure~\ref{fig:Method2-ImprovementPercentage} show the accuracy gains post-finetuning.  
While most models show only marginal improvements, suggesting their limited ability to transfer guideline knowledge to the downstream QA task, one notable exception is \texttt{Meditron-7B}, whose performance nearly doubles.  
We hypothesize two contributing factors:  
(1) \texttt{Meditron-7B} starts from a relatively low baseline, leaving greater room for improvement; and  
(2) despite being trained on medical corpora, the model may lack structured clinical reasoning skills required by MedGUIDE, which our continued pretraining helps reinforce. These results further underscore the importance of MedGUIDE—not only as a test of general medical knowledge, but as a benchmark for evaluating whether LLMs can follow structured, guideline-based clinical logic.

\section{Conclusion}\label{sec:conclusion}

In this work, we introduce \textbf{MedGUIDE}, a benchmark for evaluating the clinical reasoning abilities of LLMs grounded in standardized medical guidelines. Unlike prior benchmarks that primarily test factual recall or domain-specific knowledge, MedGUIDE emphasizes stepwise diagnostic reasoning and adherence to expert-defined clinical pathways. Through a high-quality QA dataset generated from 55 NCCN decision trees and a dual-stage quality filtering pipeline, we evaluate 25 general and medical LLMs, revealing significant gaps in existing models' ability to reason through structured decision logic. Notably, medical LLMs often underperform despite their specialized training, and guideline grounding via contextual input or continued pretraining yields limited improvements, with few exceptions. These findings suggest that beyond domain knowledge, robust guideline-following and sequential reasoning remain open challenges. We hope MedGUIDE serves as a valuable resource for driving progress toward clinically useful, safety-critical LLM applications.

\bibliographystyle{plainnat}
\bibliography{reference}


\newpage 
\appendix
\label{sec:append}

\hypersetup{linkcolor=black} 
\startcontents[appendix]
\printcontents[appendix]{ }{0}{
    \section*{Appendix}
}
\hypersetup{linkcolor=black} 

\vspace{2mm}

\clearpage

\section{More Related Works} \label{sec:Appendix-MoreRelatedWorks}

\section{Limitations and Future Directions} \label{sec:Appendix-Limitations}
\paragraph{Limitations and Future Work.} 
While MedGUIDE is a comprehensive benchmark grounded in real clinical guidelines, there are natural opportunities for extension. Our current focus is on NCCN oncology guidelines, and future versions could expand to cover other specialties such as cardiology or endocrinology to further assess cross-domain generalization. Also, we explored guideline-grounded fine-tuning and context augmentation, future work may explore reinforcement learning or retrieval-enhanced methods to further align models with structured clinical reasoning. Overall, we see MedGUIDE as a solid foundation upon which richer diagnostic benchmarks can be built.

\section{Demographic Information of Annotators}\label{sec:Appendix-Annotators}

\begin{table}[ht]
\renewcommand{\arraystretch}{1.3}  
\centering
\begin{tabular}{l p{8cm} l l}
\hline
\textbf{Annotator ID} & \textbf{School \& Program} & \textbf{Year} & \multicolumn{1}{l}{\textbf{Age}} \\ \hline
01 & Harvard Medical School & M3 & 27 \\ \hline
02 & \makecell[l]{Harvard Bioengineering \& Department of \\Stem Cell and Regenerative Biology (HSCRB)} & G5 & 27 \\ \hline
03 & \makecell[l]{Harvard Department of Stem Cell \\and Regenerative Biology (HSCRB)} & N/A  & 24 \\ \hline
04 & Harvard Population Health Science - Statistical Genetics & incoming PhD & 24  \\ \hline
05 & Harvard Population Health Science - Epidemiology & G1 & 25 \\ \hline
06 & MIT Biology & PhD & 28  \\ \hline
\end{tabular}
\end{table}

\section{Disease Abbreviations} \label{sec:Appendix-DiseaseAbbreviations}

Table~\ref{tab:disease-abbreviation} lists the abbreviations used in our dataset along with the corresponding full disease names derived from NCCN guidelines.

\begin{table}[h]
\centering
\begin{tabular}{ll}
\toprule
\textbf{Abbreviation} & \textbf{Full Disease Name} \\
\midrule
\texttt{aml} & Acute myeloid leukemia \\
\texttt{bla} & Bladder cancer \\
\texttt{bon} & Bone cancer \\
\texttt{bre} & Breast cancer \\
\texttt{cer} & Cervical cancer \\
\texttt{cml} & Chronic myeloid leukemia \\
\texttt{cns} & Central nervous system cancer \\
\texttt{col} & Colon cancer \\
\texttt{eso} & Esophageal and esophagastric junction cancer \\
\texttt{gas} & Gastric cancer \\
\texttt{hcc} & Hepatocellular carcinoma \\
\texttt{nscl} & Non-small cell lung cancer \\
\texttt{pancreatic} & Pancreatic cancer \\
\texttt{prostate} & Prostate cancer \\
\texttt{sclc} & Small cell lung cancer \\
\texttt{thyroid} & Thyroid carcinoma \\
\texttt{vulvar} & Vulvar cancer \\
\bottomrule
\end{tabular}
\caption{Abbreviations used for disease categories in the MedGUIDE benchmark.}
\label{tab:disease-abbreviation}
\end{table}

\section{Criteria for QA Data Quality Evaluation}\label{sec:Appendix-Criteria}

Here we provide detailed criteria for both clinical and general aspects, along with their corresponding scoring rubrics.

\subsection{Clinical Evaluation Criteria}

\subsubsection*{1. Clinical Plausibility}
\textbf{Definition:} Assess how realistic, accurate, and medically plausible the patient scenarios and clinical histories are.

\begin{itemize}
    \item \textbf{Score 5 (Excellent):} Entirely realistic scenario; demographics, disease course, treatment history, and outcomes accurately reflect real-world clinical practice.\\
    \textit{Example:} A 65-year-old male smoker with stage III NSCLC, weight loss, and persistent cough undergoes CT and biopsy confirming adenocarcinoma.
    \item \textbf{Score 4 (Good):} Mostly realistic with minor atypical or simplified details.\\
    \textit{Example:} 30-year-old woman with hemoptysis and imaging-confirmed large lung mass, but no risk factors mentioned.
    \item \textbf{Score 3 (Fair):} Plausible but includes inconsistencies or missing details.\\
    \textit{Example:} 75-year-old with metastatic prostate cancer but no urinary symptoms or PSA history.
    \item \textbf{Score 2 (Poor):} Multiple clinical inaccuracies.\\
    \textit{Example:} 25-year-old female with thyroid cancer metastasizing to bones without prior neck symptoms.
    \item \textbf{Score 1 (Implausible):} Medically impossible or significantly flawed.\\
    \textit{Example:} Localized thyroid cancer treated with first-line systemic chemotherapy.
\end{itemize}

\subsubsection*{2. Clinical Utility}
\textbf{Definition:} Evaluate educational relevance and practical clinical usefulness of the scenario and question.

\begin{itemize}
    \item \textbf{Score 5 (Excellent):} Highly relevant to key guideline-based decisions.\\
    \textit{Example:} Choosing between surveillance vs. surgery for low-risk thyroid cancer.
    \item \textbf{Score 4 (Good):} Relevant but less frequent or less critical decision.\\
    \textit{Example:} Choosing imaging modality for thyroid nodule staging.
    \item \textbf{Score 3 (Fair):} Basic or somewhat obvious decision.\\
    \textit{Example:} Referral for suspected thyroid nodule.
    \item \textbf{Score 2 (Poor):} Trivial scenario.\\
    \textit{Example:} Whether to evaluate a neck mass in a symptomatic adult.
    \item \textbf{Score 1 (Misleading):} Irrelevant or outdated guidance.\\
    \textit{Example:} Surgical treatment for incidental 3 mm thyroid cyst.
\end{itemize}

\subsubsection*{3. Quality of Decision Path}
\textbf{Definition:} Assess the clinical logic and fidelity to real-world guideline-based decision sequences.

\begin{itemize}
    \item \textbf{Score 5 (Excellent):} Fully guideline-consistent and logically coherent.
    \item \textbf{Score 4 (Good):} Minor deviation, otherwise accurate.
    \item \textbf{Score 3 (Fair):} Small errors but overall direction intact.
    \item \textbf{Score 2 (Poor):} Multiple inconsistencies or unclear transitions.
    \item \textbf{Score 1 (Invalid):} Contradicts guidelines or nonsensical path.
\end{itemize}

\subsubsection*{4. Alignment to Decision Path}
\textbf{Definition:} Evaluate if the patient profile matches the simulated guideline path.

\begin{itemize}
    \item \textbf{Score 5 (Excellent):} All key nodes correctly included.
    \item \textbf{Score 4 (Good):} Minor omissions, overall aligned.
    \item \textbf{Score 3 (Fair):} Multiple inaccuracies but generally reasonable.
    \item \textbf{Score 2 (Poor):} Weak alignment; many steps wrong or missing.
    \item \textbf{Score 1 (Misaligned):} Entirely mismatched from the guideline.
\end{itemize}

\subsubsection*{5. Clinical Accuracy of Correct Answer}
\textbf{Definition:} Confirm whether the correct answer is safe and guideline-aligned.

\begin{itemize}
    \item \textbf{Score 5 (Excellent):} Fully guideline-supported.\\
    \textit{Example:} RAI for iodine-avid thyroid cancer.
    \item \textbf{Score 4 (Good):} Minor ambiguity or edge-case.\\
    \textit{Example:} RAI for borderline thyroid case.
    \item \textbf{Score 3 (Fair):} Acceptable but not ideal.
    \item \textbf{Score 2 (Poor):} Technically possible but incorrect.
    \item \textbf{Score 1 (Unsafe):} Clinically incorrect or harmful.
\end{itemize}

\vspace{1em}
\subsection{General Quality Criteria}

\subsubsection*{1. Clarity and Detail Level}
\textbf{Definition:} Evaluate if the clinical scenario is clear, unambiguous, and sufficiently detailed.

\begin{itemize}
    \item \textbf{Score 5}: Clear, detailed, and precise medical language.
    \item \textbf{Score 4}: Mostly clear with minor ambiguities.
    \item \textbf{Score 3}: Some ambiguity or missing elements.
    \item \textbf{Score 2}: Lacks key clinical details.
    \item \textbf{Score 1}: Unclear or insufficient to interpret.
\end{itemize}

\subsubsection*{2. Consistency and Internal Logic}
\textbf{Definition:} Assess logical consistency of the patient narrative.

\begin{itemize}
    \item \textbf{Score 5}: Fully consistent with clear temporal logic.
    \item \textbf{Score 4}: Mostly consistent with minor issues.
    \item \textbf{Score 3}: Some contradictions.
    \item \textbf{Score 2}: Significant inconsistencies.
    \item \textbf{Score 1}: Illogical or medically impossible.
\end{itemize}

\subsubsection*{3. Safety and Toxicity}
\textbf{Definition:} Ensure that scenarios and choices are medically safe and non-harmful.

\begin{itemize}
    \item \textbf{Score 5}: Completely safe and appropriate.
    \item \textbf{Score 4}: Safe with minor issues.
    \item \textbf{Score 3}: Generally safe but questionable detail.
    \item \textbf{Score 2}: Contains safety concerns.
    \item \textbf{Score 1}: Unsafe or harmful.
\end{itemize}

\subsubsection*{4. Textual Quality and Professionalism}
\textbf{Definition:} Assess grammar, readability, and tone.

\begin{itemize}
    \item \textbf{Score 5}: Highly professional and grammatically correct.
    \item \textbf{Score 4}: Minor issues not affecting clarity.
    \item \textbf{Score 3}: Noticeable grammar/style flaws.
    \item \textbf{Score 2}: Hard to follow due to errors.
    \item \textbf{Score 1}: Poor quality and unprofessional.
\end{itemize}

\subsubsection*{5. Option Distinctiveness}
\textbf{Definition:} Evaluate whether answer choices are clearly distinct.

\begin{itemize}
    \item \textbf{Score 5}: All options clearly distinct.
    \item \textbf{Score 4}: Minor overlaps, still distinguishable.
    \item \textbf{Score 3}: Some overlap but manageable.
    \item \textbf{Score 2}: Significant overlap.
    \item \textbf{Score 1}: Indistinct or redundant choices.
\end{itemize}

\section{Reward Model Training}\label{sec:Appendix-RewardModelTraining}
For each input sample $x$, we have a ground truth reward vector $[{R_1(x), R_2(x), \dots, R_5(x)}]$ corresponding to the five evaluation dimensions. The multi-head reward model is trained to predict $[\hat{R}_1(x), \hat{R}_2(x), \dots, \hat{R}_5(x)]$. This is formulated as a multi-label regression task, where each label is a 5-dimensional real-valued score vector.

We use mean squared error (MSE) as the loss function across all heads:
\begin{equation}
\mathcal{L}(x) = \frac{1}{5} \sum_{i=1}^{5} \left( \hat{R}_i(x) - R_i(x) \right)^2
\end{equation}

For training, we use a single NVIDIA H100 80GB GPU. The model is trained for 5 epochs with a learning rate of $2 \times 10^{-5}$. Hyperparameters and the backbone model are selected via grid search over five candidates: \texttt{Llama3.2-1B-Instruct}, \texttt{Llama3.2-3B-Instruct}, \texttt{Llama3.1-8B-Instruct}, \texttt{Qwen2.5-1.5B-Instruct}, and \texttt{Qwen2.5-7B-Instruct}. The search spans learning rates {5e-5, 2e-5, 1e-5, 5e-6} and epoch counts ranging from 1 to 16. The validation MSE is 0.28 at the end of training.

\section{MedGUIDE Evaluation Results}
In Table~\ref{tab:Results-Base}, we present the detailed scores for both accuracy metrics across all 25 models evaluated on the MedGUIDE benchmark.
\begin{table}[ht]
\centering
\caption{Performance of 25 LLMs on MedGUIDE}
\begin{tabular}{l|c|c}
\hline
\textbf{Model} & \textbf{Accuracy} & \textbf{Weighted Accuracy} \\
\hline
Claude-3-5-Haiku-20241022 & 0.4935 & 0.4755 \\
Claude-3-7-Sonnet-20250219 & 0.5765 & 0.5562 \\
ClinicalCamel-70B & 0.4683 & 0.4360 \\
Deepseek-V3 & 0.5957 & 0.5770 \\
Deepseek-R1 & 0.6006 & 0.5812 \\
GPT-4.1 & 0.6439 & 0.6254 \\
GPT-4o-mini & 0.5557 & 0.5342 \\
O1 & 0.6239 & 0.6049 \\
O4-mini & 0.6075 & 0.5890 \\
Llama-3.1-8B-Instruct & 0.4955 & 0.4706 \\
Llama-3.1-70B-Instruct & 0.5654 & 0.5431 \\
Llama-3.2-1B-Instruct & 0.3121 & 0.2889 \\
Llama-3.2-3B-Instruct & 0.4377 & 0.4174 \\
Medalpaca-7b & 0.4116 & 0.3899 \\
Medalpaca-13b & 0.2721 & 0.2516 \\
Meditron-7b & 0.1701 & 0.1458 \\
Meditron-70b & 0.2530 & 0.2295 \\
Mistral-7B-Instruct-v0.3 & 0.5034 & 0.4813 \\
Mixtral-8x7B-Instruct-v0.1 & 0.5095 & 0.4915 \\
Qwen2.5-1.5B-Instruct & 0.4242 & 0.4044 \\
Qwen2.5-7B-Instruct & 0.5188 & 0.5007 \\
Qwen3-4B & 0.4789 & 0.4624 \\
Qwen3-8B & 0.4936 & 0.4733 \\
Qwen3-14B & 0.5554 & 0.5354 \\
Qwen3-32B & 0.5553 & 0.5366 \\
\hline
\end{tabular}
\label{tab:Results-Base}
\end{table}

\section{Correlation Between Benchmarks}\label{sec:Appendix-BenchmarkCorrelation}

\subsection{Details of Benchmarks}\label{subsec:Appendix-BenchmarkCorrelation-Details}

\paragraph{IFEval.}
\textit{Instruction-Following Evaluation (IFEval)}~\citep{zhou2023instruction}
is a synthetic benchmark designed to probe an LLM’s ability to follow
\emph{explicit, rule-verifiable constraints}.
Each of the 541 prompts is paired with one to three
constraints drawn from a catalogue of 25 rule types, covering
simple length requirements (``write at least \texttt{N} words’’),
formatting rules (``output exactly two bullet points’’),
keyword constraints (``mention the string \texttt{AI} three times’’),
and lexical bans (``do not use the word \texttt{because}’’), among others.
Because every constraint can be checked automatically,
evaluation is performed with \emph{instruction-level loose accuracy}:
a response is scored as correct if \textbf{all} constraints attached to that
prompt are satisfied, allowing for innocuous preambles or markdown wrappers
that do not violate any rule. (We follow the “loose’’ variant proposed by~\citet{zhou2023instruction}, which strips leading markdown headings and trailing boiler-plate before rule checking so as to avoid false negatives.)
In the correlation study we treat each model’s overall pass rate
(\%) on IFEval as one of the two score vectors.

\paragraph{MMLU--\textsc{professional-medicine}.}
\textit{Massive Multitask Language Understanding (MMLU)} is a 57-subject
multiple-choice benchmark that spans high-school, university, and
professional-licensing curricula~\citep{hendrycks2020measuring}.
The \textsc{professional-medicine} subset consists of
132 questions drawn from USMLE-style practice banks
and medical-board preparatory materials.
Questions demand factual recall of physiology, pathology and pharmacology
as well as short clinical reasoning:
each item provides four answer choices with exactly one correct option.
Following the MMLU protocol, we report
\emph{five-shot accuracy}: every test question is preceded by
five exemplars from the same sub-domain, and a model’s answer
is selected by its highest-log-probability choice.
This accuracy forms the second score vector in our analysis.

\paragraph{Why these two benchmarks?}
IFEval isolates pure \emph{instruction adherence} under
transparent, rule-based constraints, while
MMLU--\textsc{professional-medicine} measures
\emph{domain knowledge and clinical reasoning}.
By correlating MedGUIDE results with these two complementary axes
we can disentangle whether strong guideline reasoning
co-varies more with generic instruction-following skills,
with medical knowledge, or with both.

\subsection{Correlation Metrics for Benchmark Comparison}\label{subsec:Appendix-BenchmarkCorrelation-Metrics}

To quantify the relationship between model performance across different benchmarks, we use the following statistical correlation metrics:

\paragraph{Spearman’s Rank Correlation Coefficient.}  
Spearman’s correlation measures the strength of a monotonic relationship between two sets of scores. It is defined as the Pearson correlation between the rank-transformed variables:
\[
  \rho_S = \frac{\sum_{i=1}^n (\mathrm{rank}(x_i) - \overline{r}_x)(\mathrm{rank}(y_i) - \overline{r}_y)}{\sqrt{\sum_{i=1}^n (\mathrm{rank}(x_i) - \overline{r}_x)^2} \sqrt{\sum_{i=1}^n (\mathrm{rank}(y_i) - \overline{r}_y)^2}},
\]
where \( \mathrm{rank}(x_i) \) and \( \mathrm{rank}(y_i) \) are the ranks of \( x_i \) and \( y_i \), and \( \overline{r}_x \), \( \overline{r}_y \) are their mean ranks.

\paragraph{Kendall’s \( \tau \).}  
Kendall’s \( \tau \) measures ordinal association by counting the number of concordant and discordant pairs:
\[
  \tau = \frac{C - D}{C + D},
\]
where \( C \) is the number of concordant pairs and \( D \) is the number of discordant pairs among all \( \binom{n}{2} \) possible pairs.

\paragraph{Pearson Correlation Coefficient.}  
Pearson’s correlation measures the strength of a linear relationship between two numerical variables:
\[
  r = \frac{\sum_{i=1}^n (x_i - \bar{x})(y_i - \bar{y})}{\sqrt{\sum_{i=1}^n (x_i - \bar{x})^2} \sqrt{\sum_{i=1}^n (y_i - \bar{y})^2}},
\]
where \( \bar{x} \) and \( \bar{y} \) are the means of \( x \) and \( y \), respectively.

These metrics allow us to assess whether strong performance on MedGUIDE correlates with performance on other medical reasoning benchmarks, either in terms of rank agreement or absolute score similarity.

\paragraph{Comparison of Metrics.}  
Each correlation metric offers distinct interpretability and sensitivity characteristics. Below we summarize when each is most appropriate for comparing benchmark scores:

\begin{itemize}
    \item \textbf{Pearson correlation ($r$)} asks: \textit{Are the two score vectors linearly related?} It assumes interval-scale scores and is sensitive to outliers. This metric is most informative when the relationship between benchmarks is expected to be approximately linear (e.g., every 10-point gain on Benchmark A corresponds to an 8-point gain on Benchmark B).

    \item \textbf{Spearman correlation ($\rho$)} asks: \textit{Do higher scores on one benchmark generally correspond to higher scores on the other?} It converts scores into ranks and measures monotonic relationships. This rank-based approach is robust to non-linear transformations and outliers, making it suitable when benchmarks differ in scale or exhibit curved relationships.

    \item \textbf{Kendall's $\tau$} asks: \textit{For each pair of models, does the higher-scoring model on one benchmark also rank higher on the other?} This purely ordinal metric counts concordant and discordant pairs, making it especially robust to noise and ties. It is most informative when $N$ is small (as in our case with $N=25$) and when a probabilistic interpretation of ranking agreement is desired.
\end{itemize}

\noindent
By jointly considering these three metrics, we obtain a more comprehensive view of how two benchmarks align—capturing linear trends (Pearson), monotonicity (Spearman), and ordinal consistency (Kendall).

\subsection{Correlation Results}\label{subsec:Appendix-BenchmarkCorrelation-Results}

\begin{figure}[ht]
    \centering
    \includegraphics[width=1.0\textwidth]{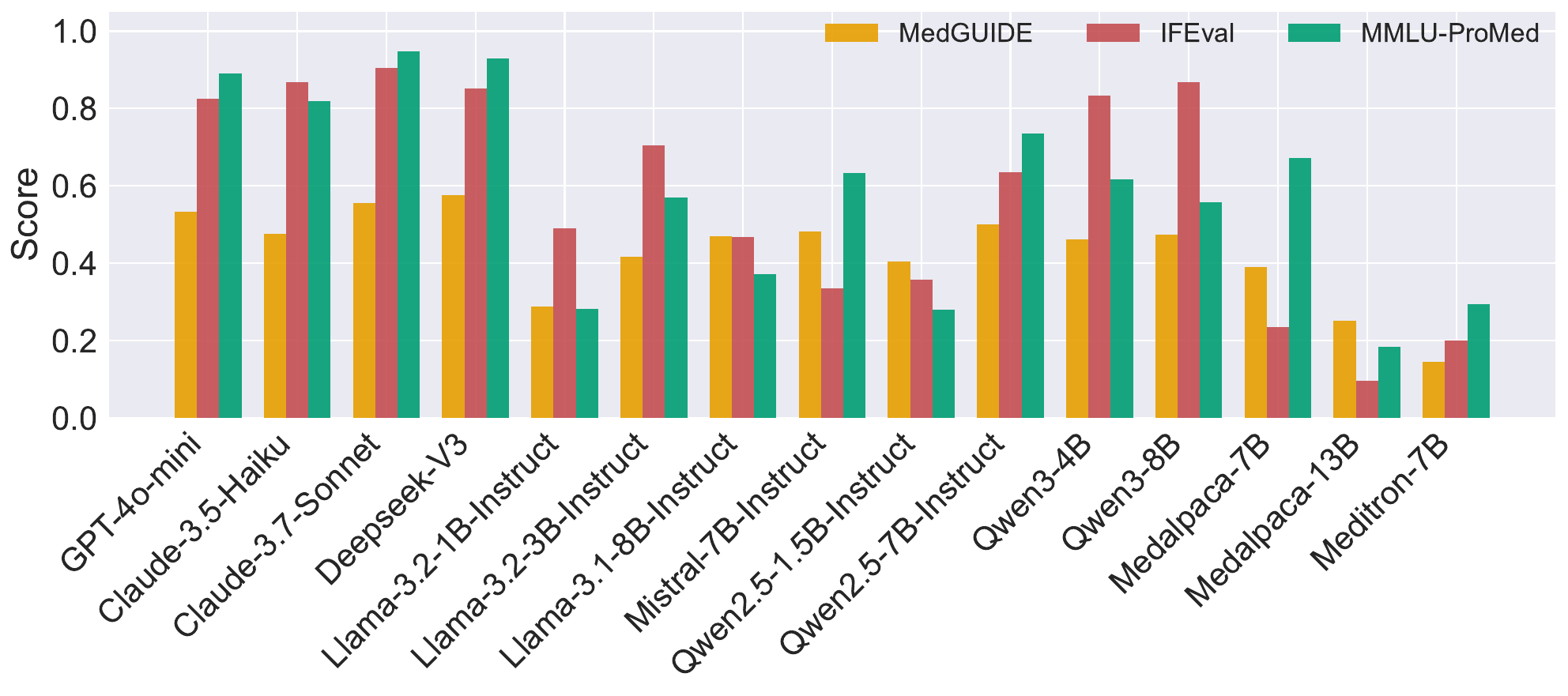}
    \caption{Performance scores of models across three benchmarks: MedGUIDE (Weighted Accuracy), IFEval, and MMLU-Professional Medicine.
}
    \label{fig:BenchmarkComparisonBarPlot}
\end{figure}

\begin{figure}[ht]
    \centering
    \includegraphics[width=0.6\textwidth]{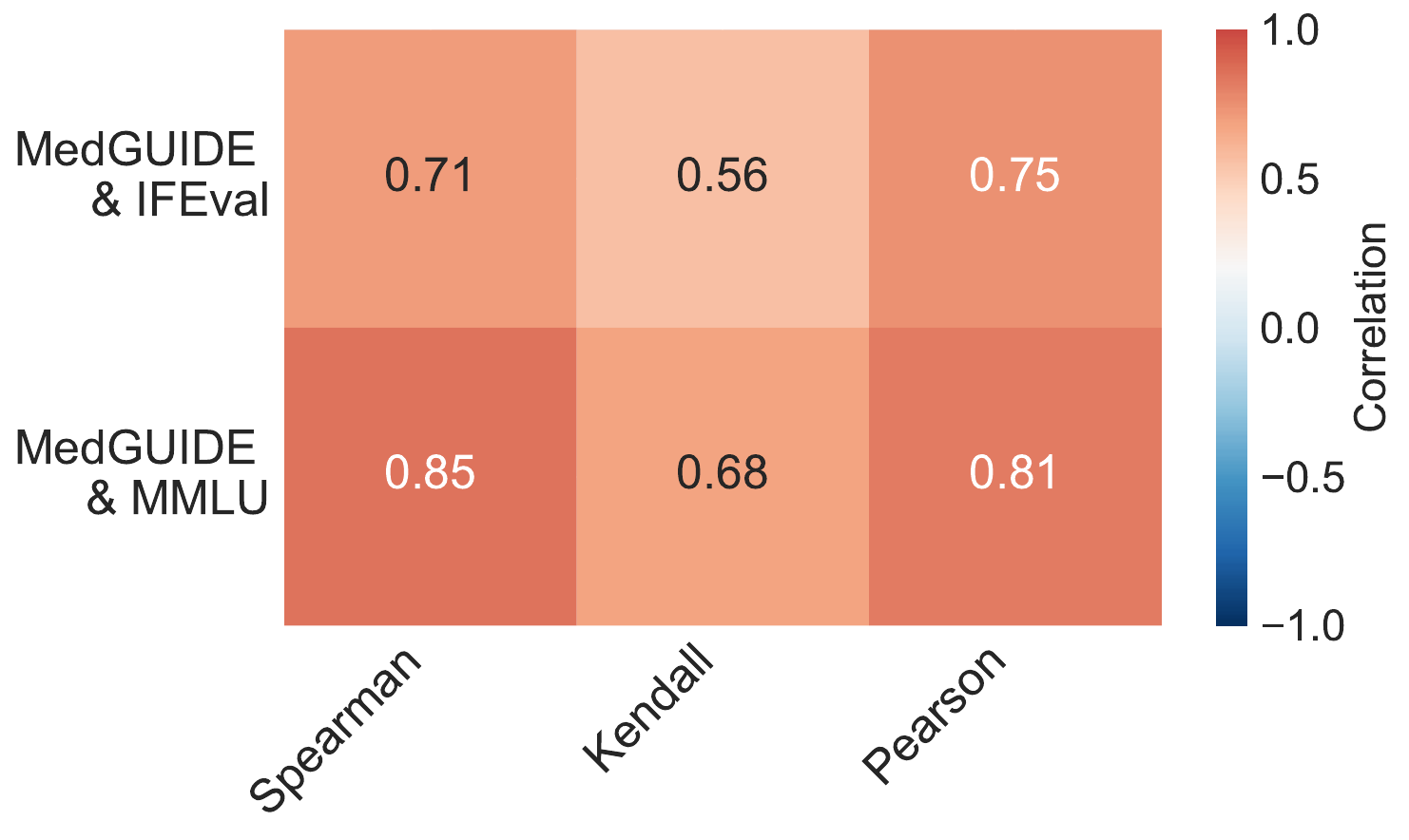}
    \caption{Pairwise correlations between MedGUIDE, IFEval, and MMLU-Professional Medicine across 25 models, evaluated via Spearman, Kendall, and Pearson metrics. MedGUIDE aligns more strongly with MMLU in absolute scoring, but shares notable ranking similarity with IFEval as well.}
    \label{fig:BenchmarkCorrelations}
\end{figure}

Figure~\ref{fig:BenchmarkComparisonBarPlot} shows the exact performance scores on three benchmarks. Figure~\ref{fig:BenchmarkCorrelations} shows correlations between \textbf{MedGUIDE} and both of \textbf{IFEval} \citep{dong2024can} and \textbf{MMLU-Professional Medicine} \citep{hendrycks2020measuring} across 25 models, using Spearman's $\rho$, Kendall's $\tau$, and Pearson's $r$ metrics.

We observe strong positive correlations between MedGUIDE and both benchmarks. Interestingly, MedGUIDE exhibits even higher alignment with MMLU (e.g., $\rho = 0.85$, $r = 0.81$) than with IFEval ($\rho = 0.71$, $r = 0.75$), suggesting that despite its emphasis on structured decision-making, MedGUIDE retains a substantial factual and knowledge-intensive component. The relatively weaker Kendall's $\tau$ values (e.g., $0.56$ for MedGUIDE–IFEval and $0.68$ for MedGUIDE–MMLU) indicate moderate agreement in relative model ranking order.

Overall, the correlations highlight that MedGUIDE partially overlaps with both instruction-following and medical knowledge evaluation, while introducing unique challenges in decision-path compliance not fully captured by either benchmark alone.

\section{Methods for Improving MedGUIDE Accuracy}\label{sec:RewardModelTraining}

\subsection{Method I}\label{subsec:Appendix-Method1}
\begin{figure}[ht]
    \centering
    \includegraphics[width=\textwidth]{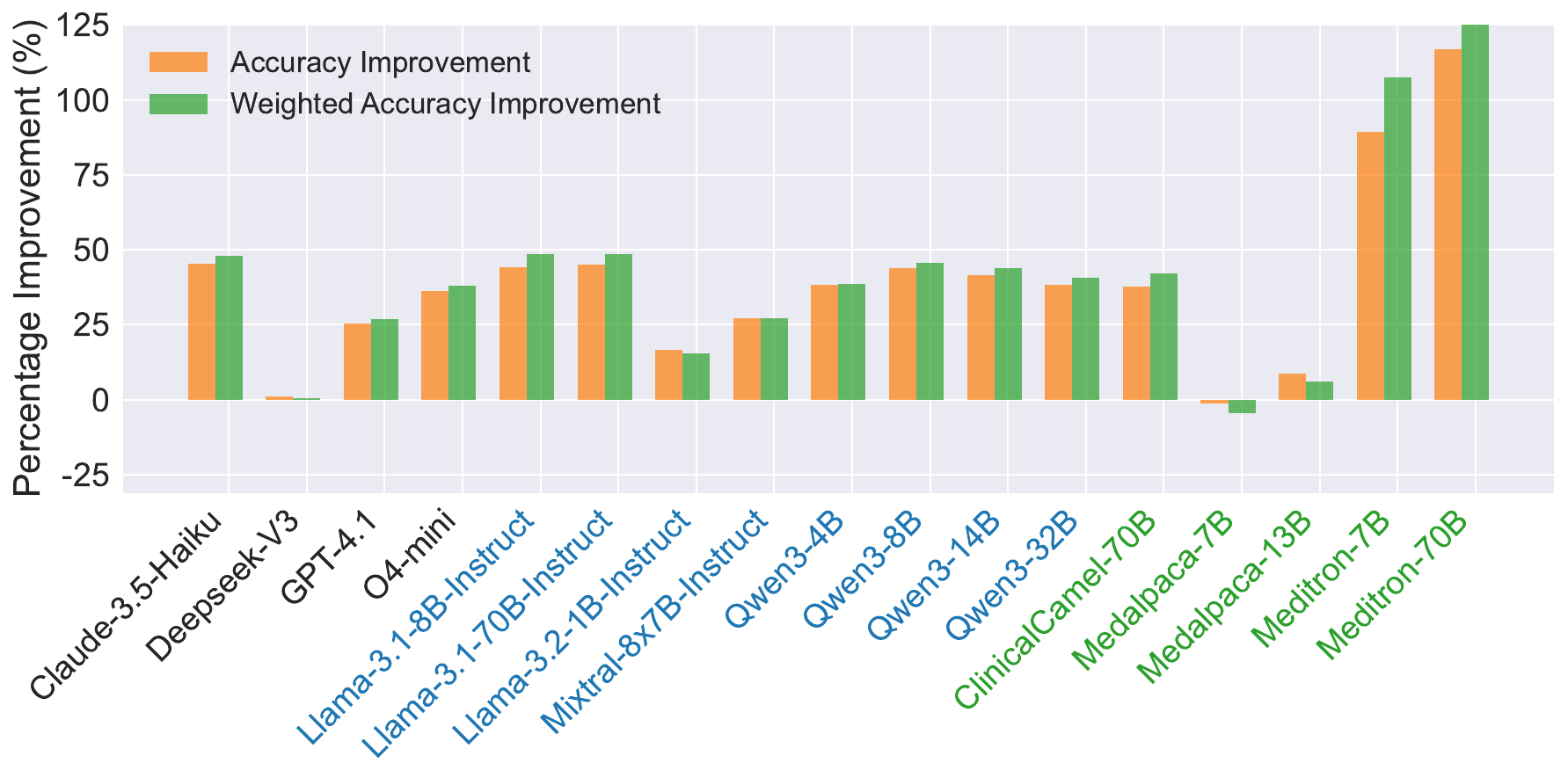}
    \caption{Improvement of accuracy and weighted accuracy for Method I}
    \label{fig:Method1-ImprovementPercentage}
\end{figure}

Below in Table~\ref{tab:Method1-ImprovementPercentage}, we report the exact accuracy scores after applying Method I (including guideline in context), along with the corresponding relative improvements in percentage. These improvements are also visualized in Figure~\ref{fig:Method1-ImprovementPercentage}. We then present a detailed case study demonstrating how Method I enables the model to follow the guideline and produce both the correct answer and a guideline-consistent explanation.

\begin{table}[htbp]
\centering
\caption{LLM performance after applying clinical guideline by Method I, with relative improvement in percentages.}
\begin{tabular}{l c c}
\toprule
Model & Accuracy & Weighted Accuracy \\
\midrule
Claude-3-5-Haiku-20241022 & 0.6725 (36\%) & 0.6579 (38\%) \\
ClinicalCamel-70B & 0.6099 (30\%) & 0.5829 (34\%) \\
Deepseek-V3 & 0.6010 (1\%) & 0.5791 (0\%) \\
GPT-4.1 & 0.7746 (20\%) & 0.7596 (21\%) \\
O4-mini & 0.7831 (29\%) & 0.7677 (30\%) \\
Llama-3.1-8B-Instruct & 0.6707 (35\%) & 0.6536 (39\%) \\
Llama-3.1-70B-Instruct & 0.7686 (36\%) & 0.7547 (39\%) \\
Llama-3.2-1B-Instruct & 0.3536 (13\%) & 0.3244 (12\%) \\
Medalpaca-7b & 0.4070 (-1\%) & 0.3755 (-4\%) \\
Medalpaca-13b & 0.2910 (7\%) & 0.2639 (5\%) \\
Meditron-7b & 0.2919 (72\%) & 0.2712 (86\%) \\
Meditron-70b & 0.4896 (94\%) & 0.4628 (102\%) \\
Mixtral-8x7B-Instruct-v0.1 & 0.6205 (22\%) & 0.5988 (22\%) \\
Qwen3-8B & 0.6668 (35\%) & 0.6465 (37\%) \\
Qwen3-4B & 0.6251 (31\%) & 0.6048 (31\%) \\
Qwen3-14B & 0.7399 (33\%) & 0.7230 (35\%) \\
Qwen3-32B & 0.7260 (31\%) & 0.7106 (32\%) \\
\bottomrule
\end{tabular}

\label{tab:Method1-ImprovementPercentage}
\end{table}

\begin{tcolorbox}[colback=blue!3!white, colframe=blue!50!white, title=Case Study of Method I, left=10mm, breakable]
\small 

\begin{center}
  \includegraphics[width=0.8\linewidth]{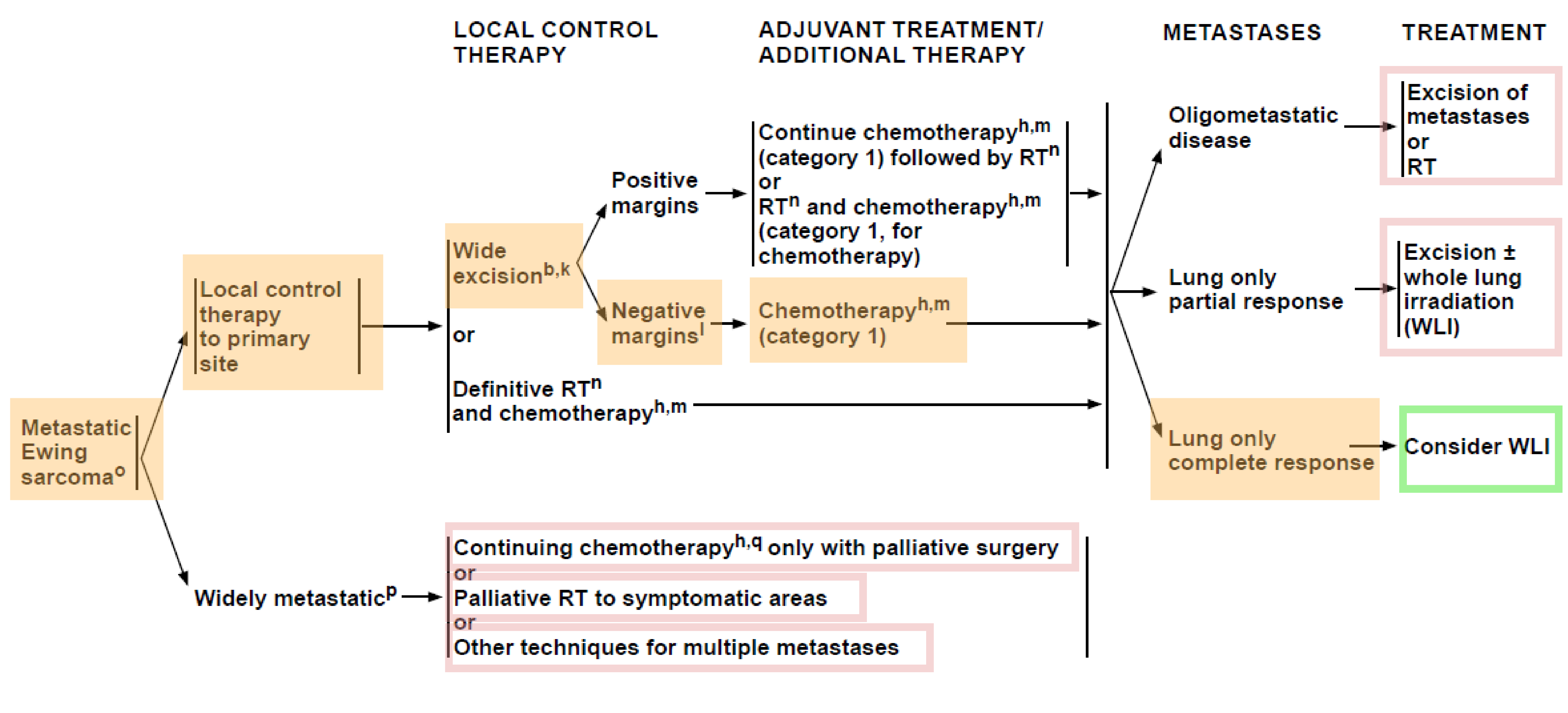}
  \captionof{figure}{\textbf{NCCN Metastatic Ewing sarcoma Guideline.} The orange boxes illustrate the workflow through which the sample QA dataset is generated. Red and green annotations represent the correct and incorrect options. }\label{fig:Appendix-ExampleDecisionTree}
\end{center}

\textbf{Prompt:}  A 16-year-old male was diagnosed with metastatic Ewing sarcoma following the evaluation of persistent pain in his right pelvis and a palpable mass. Initial imaging revealed a 5 cm tumor in the iliac wing with multiple lung metastases noted on a CT scan. He underwent neoadjuvant chemotherapy with vincristine, doxorubicin, and ifosfamide, followed by surgical resection of the primary tumor. Post-surgery, he received definitive radiation therapy to the primary site and additional chemotherapy for residual disease. Follow-up chest CT demonstrated complete resolution of the lung metastases, but there remains concern for micro-metastatic disease given the initial extensive pulmonary involvement. Given these findings, should we consider whole lung irradiation as a next step in managing his treatment?

\textbf{Options:}

\begin{enumerate}[label=(\Alph*)]
\item Excision of metastases or RT
\item Consider WLI
\item Continuing chemotherapy only with palliative surgery
\item Excision ± whole lung irradiation (WLI)
\item Palliative RT to symptomatic areas
\item Other techniques for multiple metastases
\end{enumerate}

\textbf{Correct Answer:}
B.

\bigskip \hrule \bigskip
\textbf{Original Answer by GPT4.1: \red{C}}

\textbf{Original Explanation by GPT4.1:} The correct management for this 16-year-old male with metastatic Ewing sarcoma is \red{option C: Continuing chemotherapy only with palliative surgery}. Despite the patient's comprehensive initial treatment regimen consisting of neoadjuvant chemotherapy, surgical resection of the primary iliac wing tumor, definitive radiation to the primary site, and additional chemotherapy, the history of extensive pulmonary metastases represents a significant prognostic challenge. While the follow-up chest CT demonstrates complete radiographic resolution of the visible lung metastases, patients with initial widespread metastatic Ewing sarcoma typically face a guarded prognosis. In this context, continuing systemic chemotherapy remains the backbone of treatment to control microscopic metastatic disease throughout the body, not just in the lungs. The palliative surgery component allows for addressing any local complications or symptomatic lesions that might develop, without subjecting the patient to the significant toxicity of whole lung irradiation, which carries substantial short and long-term side effects including pneumonitis, pulmonary fibrosis, and cardiac toxicity. This balanced approach prioritizes disease control while maintaining quality of life, recognizing that complete cure may be challenging in the setting of initially extensive metastatic disease. Systemic therapy offers the best chance of managing potential microscopic disease at all sites while minimizing treatment-related morbidity in this young patient who has already undergone intensive multimodal therapy.

\bigskip \hrule \bigskip
\textbf{Correct Answer by GPT 4.1 (using Method I): \teal{B}} 

\textbf{Correct Explanation by GPT 4.1 (using Method I):} To understand the decision-making path from the clinical vignette to the final option B: Consider WLI, it is essential to follow the flowchart step by step, tracing the structure logically from the root node to the leaf node. The patient is diagnosed with Metastatic Ewing Sarcoma, which represents the starting point at the root node of the flowchart. According to the guidelines, metastatic Ewing sarcoma is managed initially with Local Control Therapy to the primary site. In this case, the patient underwent surgical resection and radiation therapy, fitting the expected pathway for local control. After local control therapy, the flowchart directs the decision-making process to Adjuvant Treatment / Additional Therapy, which includes continued chemotherapy and radiation therapy. This step matches the patient's treatment history, where he received additional chemotherapy after surgery.

The next step is the Assessment of Metastatic Response, where the decision path diverges based on the response of lung metastases. The flowchart specifies three possible pathways: Oligometastatic Disease, which would prompt excision or radiation therapy; Lung Only Partial Response, suggesting incomplete control and potential excision ± whole lung irradiation (WLI); and Lung Only Complete Response. In this patient's case, imaging showed a complete response in the lung, guiding the decision towards Consider WLI as the next step to address potential micro-metastatic disease.

Following the path for Lung Only Complete Response, the guideline explicitly recommends Consider WLI for comprehensive management aimed at mitigating the risk of microscopic disease. This is where the decision path terminates, aligning precisely with the option B: Consider WLI in the clinical question. Thus, \teal{the stepwise traversal through the guideline flowchart follows a logical sequence: Metastatic Ewing Sarcoma → Local Control Therapy → Adjuvant Treatment → Lung Only Complete Response → Consider WLI, justifying B as the most clinically appropriate option}.

\bigskip \hrule \bigskip
\textbf{Analysis:} 
In this case study, we observe a clear improvement in both the reasoning and answer selection after incorporating the guideline into the prompt. In the original response (without guideline context), the model incorrectly selects option \textbf{C} and justifies its decision based on general considerations of prognosis and treatment-related toxicity. However, this explanation overlooks the patient’s complete response in the lungs and fails to recognize the appropriate next step recommended in the guideline for such cases. In contrast, after including the guideline in the prompt (Method I), the model correctly identifies \textbf{B: Consider WLI} as the recommended next step. The explanation explicitly traces the decision path through the guideline, correctly mapping the patient’s clinical course—metastatic disease, local control, adjuvant therapy, and lung-only complete response—to the corresponding recommendation. This demonstrates how incorporating the structured guideline enables the model to produce not only the correct answer but also a faithful and guideline-consistent rationale.

\end{tcolorbox}

\subsection{Method II}\label{subsec:Appendix-Method2}

\begin{figure}[ht]
    \centering
    \includegraphics[width=0.7\textwidth]{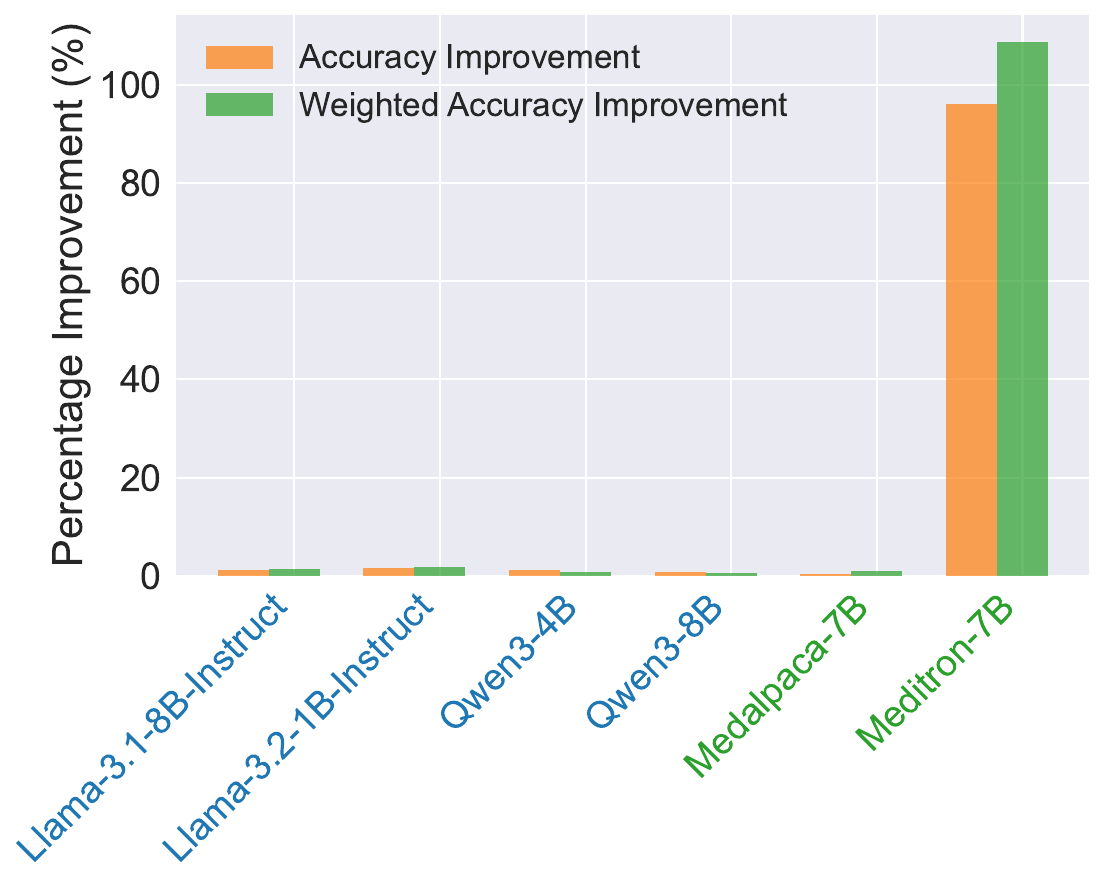}
    \caption{Improvement of accuracy and weighted accuracy for Method II}
    \label{fig:Method2-ImprovementPercentage}
\end{figure}

Below in Table~\ref{tab:Method2-ImprovementPercentage}, we report the exact accuracy scores after applying Method II (fine-tuning with guidelines), along with the corresponding relative improvements in percentage. These improvements are also visualized in Figure~\ref{fig:Method2-ImprovementPercentage}.

\begin{table}[ht]
\centering
\begin{tabular}{l c c}
\toprule
Model & Accuracy & Weighted Acc \\
\midrule
Llama-3.1-8B-Instruct & 0.5014 (1\%) & 0.4772 (1\%) \\
Llama-3.2-1B-Instruct & 0.3170 (2\%) & 0.2941 (2\%) \\
Medalpaca-7b & 0.4135 (0\%) & 0.3936 (1\%) \\
Meditron-7b & 0.3334 (96\%) & 0.3044 (109\%) \\
Qwen3-4B & 0.4844 (1\%) & 0.4661 (1\%) \\
Qwen3-8B & 0.4972 (1\%) & 0.4757 (1\%) \\
\bottomrule
\end{tabular}
\vspace{0.5em}
\caption{Performance of LLMs after fine-tuning by Method II, with improvements relative to baseline.}
\label{tab:Method2-ImprovementPercentage}
\end{table}

\section{Prompt Templates for Guideline-to-QA Generation}
\label{sec:Appendix-prompts}

To generate high-quality medical QA data from NCCN guideline figures, we design a set of structured prompts to guide large language models through three stages: (1) converting clinical guideline diagrams into structured JSON; (2) extracting all clinical decision paths from the tree; and (3) generating realistic patient vignettes and clinical questions based on each decision path. Below we provide a description and the exact text of each prompt used in this process.

\subsection{Prompt: Convert Guideline Diagram to JSON Tree}

This prompt asks the LLM to read a clinical decision tree (e.g., a screenshot of an NCCN guideline) and convert it into a machine-readable, hierarchically structured JSON format.

\begin{quote}
\ttfamily
You are given a clinical decision tree diagram (as a screenshot image) that outlines medical guidelines for diagnosis or treatment. Your task is to convert the decision tree into a structured JSON format that precisely mirrors the full logic and hierarchy presented in the figure.

* The JSON must accurately reflect all branching paths, decision points, conditions, and outcomes.\\
* All treatment options, diagnostic steps, and relevant notes (e.g. footnotes or eligibility criteria) must be preserved.\\
* Maintain the exact hierarchical relationships and nesting, so that the JSON could be used to programmatically reconstruct the original tree.\\
* Use clear and descriptive key names based on the text in the image.\\
* If treatments or conditions include multiple options or logical conditions (e.g. “A or B”), represent them using lists or nested structures as appropriate.

Example format: \\
\{\\
\quad "First relapse (morphologic or molecular)": \{\\
\quad\quad "Early relapse (<6 mo) after ATRA and arsenic trioxide (no anthracycline)": \{\\
\quad\quad\quad "Therapy": [\\
\quad\quad\quad\quad "Anthracycline-based regimen as per APL-3",\\
\quad\quad\quad\quad "Gemtuzumab ozogamicin"\\
\quad\quad\quad ]\\
\quad\quad\},\\
\quad\quad "Late relapse ($\geq$6 mo) after arsenic trioxide-containing regimen": \{\\
\quad\quad\quad "Therapy": [\\
\quad\quad\quad\quad "Arsenic trioxide ± ATRA ± (anthracycline or gemtuzumab ozogamicin)"\\
\quad\quad\quad ]\\
\quad\quad\}\\
\quad\},\\
\quad "No remission": \{\\
\quad\quad "Next steps": [\\
\quad\quad\quad "Clinical trial",\\
\quad\quad\quad "Matched sibling or alternative donor HCT"\\
\quad\quad ]\\
\quad\}\\
\}
\end{quote}

\subsection{Prompt: Extract All Decision Paths}

This prompt asks the LLM to enumerate every possible decision path from the root to a leaf node in the structured guideline.

\begin{quote}
\ttfamily
You are provided with an image of an NCCN clinical decision tree guideline, outlining detailed medical instructions for diagnosis and treatment. Your task is to precisely list all possible clinical decision paths depicted in the guideline.

Instructions: \\
* Represent each decision path as a Python list of strings.\\
* Each string must exactly match the text appearing in the decision nodes, conditions, or treatment steps from the image, without abbreviation, modification, or paraphrasing.\\
* Include every potential pathway from the initial decision node down to each final leaf node.

Now, generate all possible paths.
\end{quote}

\subsection{Prompt: Generate a QA Sample from a Decision Path}

This prompt instructs the model to generate a realistic patient vignette and a clinical question that follows the logic of a specific decision path.

\begin{quote}
\ttfamily
You are provided with a clinical decision path derived from an NCCN guideline, presented as an ordered list of decision nodes (each node is a string). Your task is to generate a realistic patient vignette—a brief clinical case—including pertinent medical history, timing of relapse, previous treatments, test results, and clinical assessments required to match precisely each node in the provided decision path from root to leaf. Conclude your vignette with a clinical question asking explicitly about the appropriate next treatment step. The correct answer must correspond exactly to the final node of the provided path but should not be mentioned explicitly in either the vignette or the question.

\medskip
Decision Path:
\begin{itemize}
  \item "First relapse (morphologic or molecular)"
  \item "Early relapse (<6 mo) after ATRA and arsenic trioxide (no anthracycline)"
  \item "Therapy"
\end{itemize}

\medskip
Formatting Instructions:
\begin{itemize}
  \item Present the entire vignette and concluding question as a single paragraph.
  \item Do not reveal or imply the correct (leaf node) answer within the vignette or the question.
\end{itemize}
\end{quote}


\end{document}